\documentclass[10pt,journal,compsoc]{IEEEtran}
\IEEEoverridecommandlockouts
\usepackage[%
  square,        
  comma,         
  numbers,       
  sort&compress 
]{natbib}

\usepackage{amsmath,amssymb,amsfonts}
\usepackage{algorithmicx}
\usepackage{algorithm} 
\usepackage{algpseudocode}
\usepackage{graphicx}
\usepackage{multirow}
\usepackage{textcomp}
\usepackage{xcolor}
\usepackage{url}
\usepackage{color}
\usepackage{booktabs}
\usepackage{hyperref}
\hypersetup{colorlinks,citecolor=blue,urlcolor=blue}
\usepackage{xcolor}
\usepackage{tikz}
\usepackage{amssymb}
\usepackage{array}
\usepackage[export]{adjustbox}
\usepackage[column=0]{cellspace}
\usepackage{pifont}
\usepackage[switch]{lineno}
\usepackage{graphviz}

\begin{document}
%
\title{RenAIssance: A Survey into AI Text-to-Image Generation in the Era of Large Model}

\author{
Fengxiang~Bie,
Yibo~Yang, 
Zhongzhu Zhou, 
Adam Ghanem,
Minjia Zhang,
Zhewei Yao,
Xiaoxia Wu,
Connor Holmes,
Pareesa Golnari,
David A. Clifton,
Yuxiong He,
Dacheng Tao, 
Shuaiwen Leon Song
 \IEEEcompsocitemizethanks{\IEEEcompsocthanksitem F. Bie, Z. Zhou, A. Gahem, D. Tao, S L. Song are with the University of Sydney, Australia.
\IEEEcompsocthanksitem Y. Yang is wth King Abdullah University of Science and Technology.
\IEEEcompsocthanksitem M. Zhang, Z. Yao, X. Wu, C. Holmes, P. Golnari, Y. He and S L. Song are with Microsoft, US.
\IEEEcompsocthanksitem D.A. Clifton is with the University of Oxford, UK.
}
}

\IEEEtitleabstractindextext{%
\begin{abstract}
Text-to-image generation (TTI) refers to the usage of models that could process text input and generate high fidelity images based on text descriptions. Text-to-image generation using neural networks could be traced back to the emergence of Generative Adversial Network (GAN), followed by the autoregressive Transformer. Diffusion models are one prominent type of generative model used for the generation of images through the systematic introduction of noises with repeating steps. As an effect of the impressive results of diffusion models on image synthesis, it has been cemented as the major image decoder used by text-to-image models and brought text-to-image generation to the forefront of machine-learning (ML) research. In the era of large models, scaling up model size and the integration with large language models have further improved the performance of TTI models, resulting the generation result nearly indistinguishable from real-world images, revolutionizing the way we retrieval images. Our explorative study has incentivised us to think that there are further ways of scaling text-to-image models with the combination of innovative model architectures and prediction enhancement techniques. We have divided the work of this survey into five main sections wherein we detail the frameworks of major literature in order to delve into the different types of text-to-image generation methods. Following this we provide a detailed comparison and critique of these methods and offer possible pathways of improvement for future work. In the future work, we argue that TTI development could yield impressive productivity improvements for creation, particularly in the context of the AIGC era, and could be extended to more complex tasks such as video generation and 3D generation.
\end{abstract}

\begin{IEEEkeywords}
Text-to-Image, Diffusion Model, DPM, CLIP, Language Model, Vision Transformer, VAE, GAN, Large model
\end{IEEEkeywords}}

\maketitle
\IEEEdisplaynontitleabstractindextext
\IEEEpeerreviewmaketitle

\ifCLASSOPTIONcompsoc
\IEEEraisesectionheading{\section{Introduction}\label{sec:introduction}}
\else
\section{Introduction}
\label{sec:introduction}
\fi
\IEEEPARstart{I}{n} recent years, AI Generated Content (AIGC) has emerged as a powerful tool for content creation, which has the potential to revolutionize the way we retrieve and consume information. AIGC refers to the use of artificial intelligence to create written, visual, or audio contents for creative, educational, and a plethora of other application cases. With advancements in the field of natural language processing (NLP) and computer vision, AIGC is capable of producing high-quality content that is progressively becoming indistinguishable from the content generated by human writers, designers, and artists. For instance, language generation models like ChatGPT \cite{chatgpt-3,chatgpt4} demonstrate a remarkable ability to understand complex language and respond with expertly crafted text, closely aligned with the input query. At the same time, image generation models   \cite{stable_diffusion,Dalle2,Parti,fontgeneration,cycle_gan,semantic_generation,inversionbased_imagetranslation,plugandplay,palette,repaint,blended_diffusion,paint_by_example} have made remarkable strides in multiple tasks, rendering AI-generated images that rival those created by human designers or real-world photographs.

Building upon this foundation, it becomes evident that applying NLP and computer vision models to AIGC holds great significance. Researchers have explored the synergistic potential of these models, seeking to leverage knowledge from both visual and textual domains, which generate visually coherent and contextually relevant images based on textual descriptions. With the combination of generative models   \cite{VAE_Inference,GAN_Model,DDPM,ViT,U-net} and large language models   \cite{Transformer,T5,Bert,GPT-2}, it is now possible to automatically generate high-fidelity images that closely match complex text descriptions. This advancement opens up new possibilities for creative expression, design, and multimedia content creation. The era of large models has facilitated the scaling of model size and training datasets, resulting in further enhanced performance for text-to-image (TTI) generation models. Consequently, the gap between generated images and those produced by top artists and photographers is diminishing. TTI generation has made a profound impact across a wide range of fields and industries, boosting productivity and saving substantial time in art design. As a result, TTI generation has become one of the most hotly debated and researched topics within the realm of AIGC. Nowadays, any individuals or industrial professionals can easily retrieve AI generated images through on-device models or online flatforms, such as Stable Diffusion \cite{stablediffusionweb}, PARTI \cite{Parti}, Dreambooth \cite{dreambooth}, NovelAi \cite{novelai}, and Midjourney \cite{midjourney}.
\begin{figure*}[htp] \centering{
\includegraphics[scale=0.58]{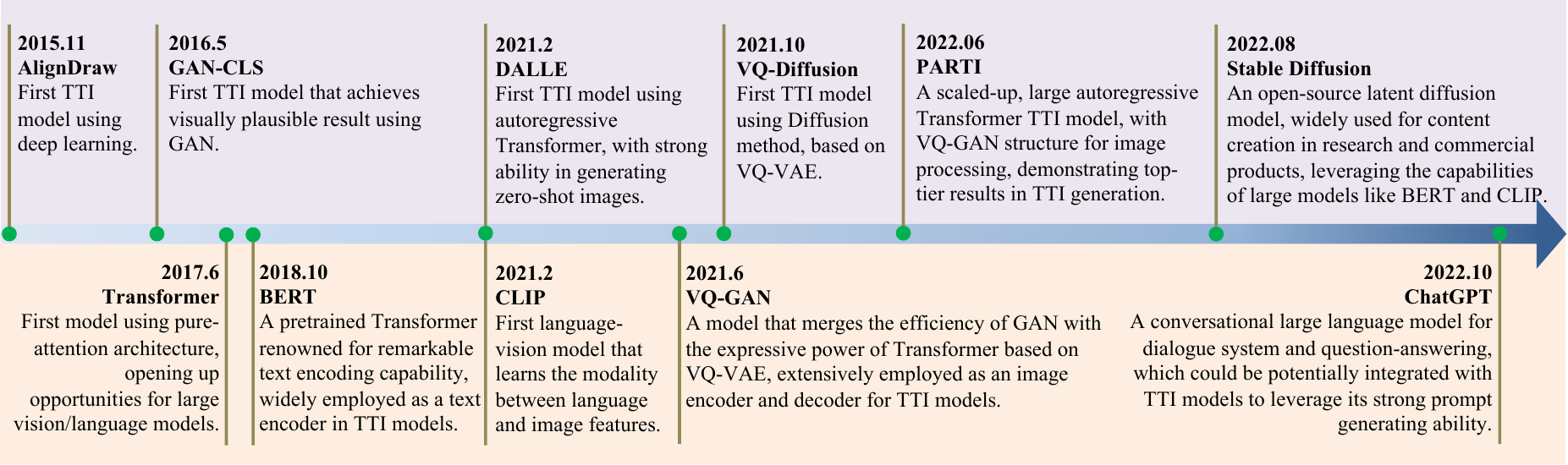}}
\caption{The milestones of text-to-image (TTI) models and large models. The upper part in light purple shows the key TTI models with high impacts, and the lower part in light yellow shows the progress of large models that provide the stimuli for the development of TTI models.}
\label{fig:milestone}
\end{figure*}

The field of TTI models has witnessed significant development, evolving from simple image generation capabilities \cite{AlignDraw,Gan-CLS} to the impressive capacity to create complex and realistic images \cite{Imagen,stable_diffusion,muse}.  Researchers have explored various deep learning \cite{Deeplearning} architectures, in order to improve image quality and sampling efficiency. During the early stages of TTI generation, generative adversarial network (GAN) \cite{GAN_Model} served as the primary backbone generative model \cite{Gan-CLS,AttentionGAN,StyleGan,DMGAN,DCGAN,MirrorGan,spatial_aware_gan}, which generates images efficiently with a generator-discriminator structure \cite{GAN_Model}. Another novel approach for TTI generation is the Autoregressive model \cite{DALLE,Parti,muse}, which employs the Transformer \cite{Transformer} architecture to predict image tokens sequentially based on text tokens. More recently, there is a clear trend that Diffusion model has gained popularity as a prominent method \cite{stable_diffusion,Dalle2,Imagen,Glide}. This approach denoises images by modeling them from Gaussian distributions \cite{do2008multivariate} using a Markovian-chain process \cite{DDPM,DDPM_Beats_Gan}. The ongoing innovations in architecture improvements and augmentation techniques have substantially enhanced Diffusion TTI models, enabling them to generate higher fidelity and more diverse images. 
 
Each major type of TTI model has its strengths and weaknesses. GAN models are efficient in scale and are able to generate images in a negligible amount of time. However, they may suffer from fidelity loss and lack intricate image details compared to autoregressive Transformer and Diffusion models \cite{DDPM_Beats_Gan}. Autoregressive Transformer has the ability to generate images highly realistic and comparable to the performance of the Diffusion method. The drawback is that they usually consists of more than 5B  \cite{DALLE,Parti,cogview2} parameters, which is much larger in model size and makes it hard to deploy on a wide range of local devices. Autoregressive Transformer predicts image tokens in a non-parallel way, which impacts the efficiency of the inference process. Despite the fact that diffusion TTI models have a reasonable model size and also achieve top results in TTI generation, it requires 50-200 sampling steps which significantly prolongs total time consumption during inference. While early GAN models may have lagged behind state-of-the-art autoregressive and diffusion models, recent advancements, combined with the benefits of large language models, have significantly narrowed the gap \cite{lafite, galip, GigaGan}. This progress has fostered improved performance in GAN models, making them more competitive in comparison to new autoregressive and diffusion models.  Given the unique advantages of each model architecture, such as model size, time efficiency, and their demonstrated capability to generate high-fidelity and diverse images, researchers continue to explore and push the boundaries of all three types of TTI models. The accomplishments of various large language models have also significantly catalyzed the advancement of TTI models. These models serve as the bedrock for robust text encoders and exemplify the evolution of more sophisticated model architectures, as illustrated in Fig \ref{fig:milestone}.

As all types of TTI models have experienced considerable improvements, it becomes evident that no single model type holds an absolute advantage over others. Therefore, a comprehensive survey that encompasses recent innovations and advancements in all TTI models becomes necessary to facilitate a clear understanding of their development and enable researchers to conveniently compare different model. To our best knowledge, previous surveys have focused on text-to-image generation only on a certain type of generative models, such as surveys only for GAN TTI generation \cite{GAN_survey1,GAN_survey2,GAN_survey3,GAN_Survey4}, and surveys focusing on diffusion models   \cite{Diffusion_survey1,diffusion_survey2}, we are the first survey that gives a holistic overview connecting all types of TTI models, and consider their development under the influence of large models. By addressing this knowledge gap and aligning our content with the current trends in deep learning, our survey serves as a crucial resource in the landscape of TTI research. In summary, we recognize the key contributions of our survey as follows:
\begin{enumerate}
    \item We give a detailed introduction to the vital components of TTI models, including different generative models, language models, and vision models.
    \item Our survey encompasses multiple types of text-to-image generative models, not only those developed in the early stages, but also modern models that have been influenced by the advancements of large models. This inclusive perspective allows us to capture the full spectrum of TTI model development.
    \item We compare the performance of different types of TTI models with both visual results and statistical results. We further discuss pros and cons of each model based on the comparison results.
    \item We outlook the potential development of TTI models, foreseeing their applications to industries and different research aspects. 
\end{enumerate}

\section{Background}
\begin{figure}[htp] \centering{
\includegraphics[scale=0.35]{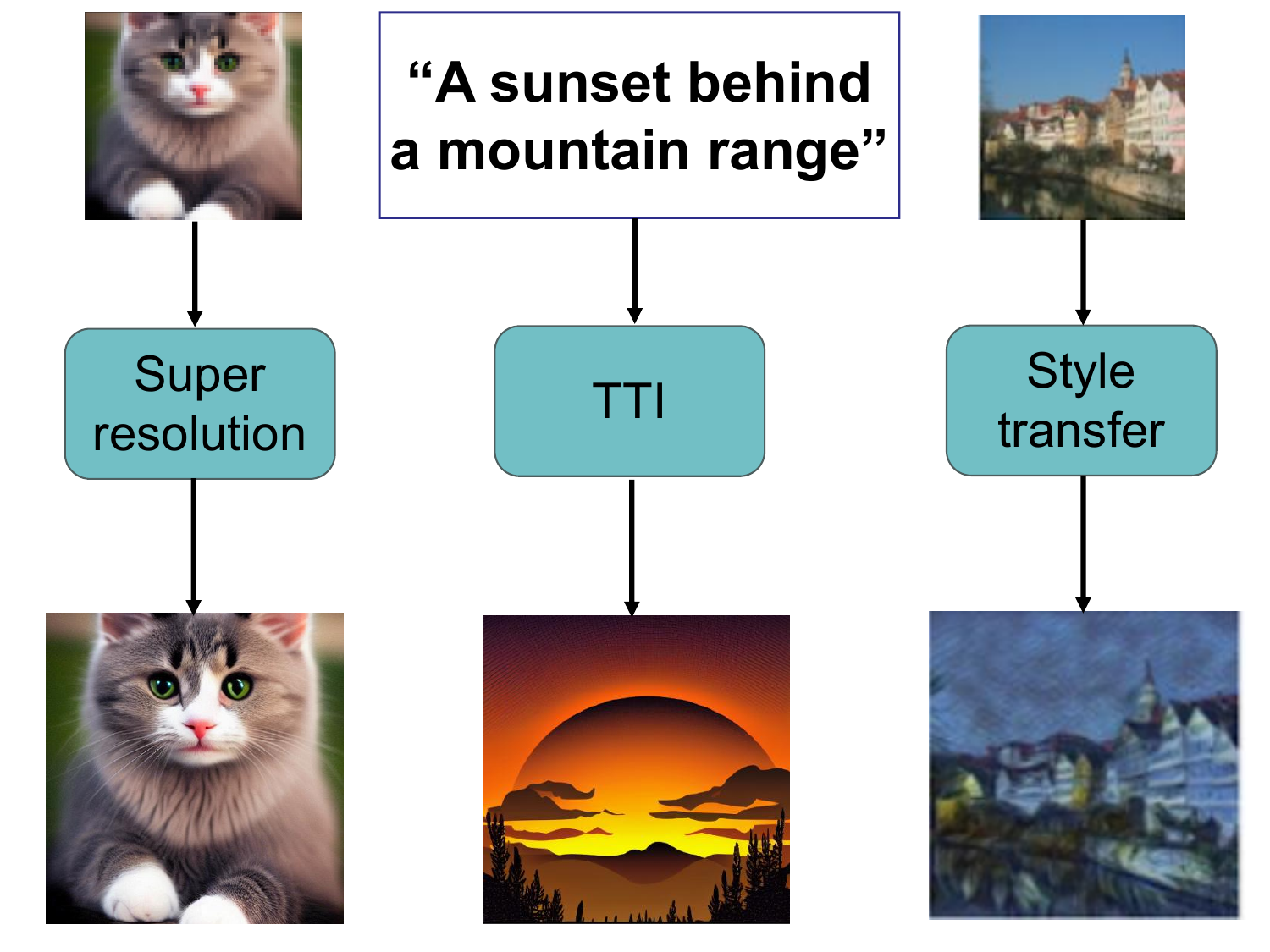}}
\caption{Examples of common tasks for image generation, including super-resolution (left), text-to-image generation (middle) and style change (right). Super-resolution enhances the image by adding more details and generates the output with high fidelity. Style change transfers the image to another domain with mutual information. Text-to-image generates high quality images that are well aligned with user input.\label{fig:common-image-tasks}}
\end{figure}
\subsection{AI Image generation}
AI image generation involves the use of neural networks to generate new images that resemble a set of input images or even have the ability to manufacture creative pieces of work. Image generation is always a heated topic under the computer vision area. Popularization of the convolutional neural network \cite{CNN, wu2017introduction} and Transformer models \cite{Transformer} has pushed the success in many works such as image-segmentation \cite{segdiff,SEGAN,wu2023towards,yang2019sognet}, super-resolution \cite{deeplearning_SR,pyramid_gan_superresolution,srdiff, lu2022transformer}, style change \cite{palette,repaint,blended_diffusion,paint_by_example}, object detection \cite{li2023transformer,zhou2022transvod,yuan2022polyphonicformer} and text-to-image generation. Fig \ref{fig:common-image-tasks} shows examples for popular image generation tasks. Super-resolution model generates images from low resolution to a higher resolution space. Unlike traditional methods such as bi-linear interpolation that results in a blurred image, super-resolution, through the use of an additional neural network, adds new details to the image and results in a high fidelity for the output image. Style transfer \cite{cycle_gan} using deep learning could transfer input images to another domain but keeps most of the semantic meaning. Text-to-image generation takes a text sentence and generates a variety of images that are all well-aligned with the text description. Text-to-image generation has close connections to daily lives and has the potential to be applied to a wide range of research and industries. Therefore, it has become one of the most discussed topics in image generation. However, text-to-image generation is hard to implement, given that it involves the complex connection of human language and visual representation, which are two separate fields in computer science research. Multiple tools and models are required to support each process of TTI generation pipeline. For example, language models are necessary for processing text inputs as conditional information to guide image generation. Vision models may also be required for encoding image data. Generative models also work as essential components used for image generation, such as GAN and U-net. VAE models could be used to map image information to a latent space. 
\subsection{Generative models}

\subsubsection{VAE: Variational Autoencoders}

Variational autoencoders (VAE) \cite{Auto-Encoding,VAE_Inference} are a class of generative models comprising of an encoder-decoder \cite{cho2014properties} structure that is able to reconstruct input data. The encoder maps each data element $x_i$ to a latent space $z$ that follows a Gaussian distribution through the generation of a mean and variance within the latent space:

\begin{equation}
\log q_{\phi}(z^{(i)}|x^{(i)}) = \log \mathcal{N}(z^{(i)} ;\mu^{(i)},\sigma^{2(i)}I).
\end{equation}

The variance of the Gaussian distribution is critical in providing the generative quality as noises are incrementally introduced to the latent space. The encoder learns to predict the variance and mean of the distribution, however, this process here is not differentiable as we are sampling from a distribution we are parameterizing. Thus, VAE makes use of a re-parameterization trick which separates the process of sampling and the prediction of the mean and variance:

\begin{equation}
z_i=\mu_i + \epsilon \times \sigma_i,
\end{equation}
where $\epsilon$ is the noise randomly sampled from the normal Gaussian distribution.
The decoder's general aim is to generate the output data to be as close as the given input data. This process is done by maximising the probability of generating input data given the latent space variable $P(x^i | z)$. VAE training is based on maximizing the tractable lower bound $L_b$ of $P(X)$:
\begin{equation}
    L_b = -D_{KL}(q(z|x)|p(z)) +\frac{1}{L} \sum_{i=1}^L(logP(x|z)).
\end{equation}

Since the encoder aims to regulate the latent space in the normal Gaussian distribution $P(Z|X) = \mathcal{N}(0,I)$, the loss function could be rewritten as:
\begin{equation}
\begin{split}
    \mathcal{L}(\theta,x^i)=-\frac{1}{2} \sum_{i=1}^d \Big(\mu_{(i)}^2 + \sigma_{(i)}^2 - \log \sigma_{(i)}^2 - 1\Big)\\+\frac{1}{L} \sum_{i=1}^L(logP(x^i|z)).
\end{split}
\end{equation}
Compared to autoencoders, VAE applies a probabilistic model to deep neural networks. VAE's main purpose is to generate a data distribution rather than giving an output with fixed numbers as done by autoencoders. Thus, sampling from the distribution increases the data diversity as well as the capability of noise-handling.

There are some limitations for VAE models. For image generation, directly sampling from a Gaussian distribution always results in a blurred image. Besides this, information loss occurs when projecting the data to a lower dimension. Therefore, VAE has relatively less performance for image synthesis compared to the state-of-the-art generative models such as GAN and Diffusion models. However, the latent space structure works as a useful feature presentation in text-to-image tasks, and it can be used for significant computation reduction combined with diffusion models  \cite{stable_diffusion, VQ-diffusion}.
\begin{figure*}[htp] \centering{
\includegraphics[scale=0.55]{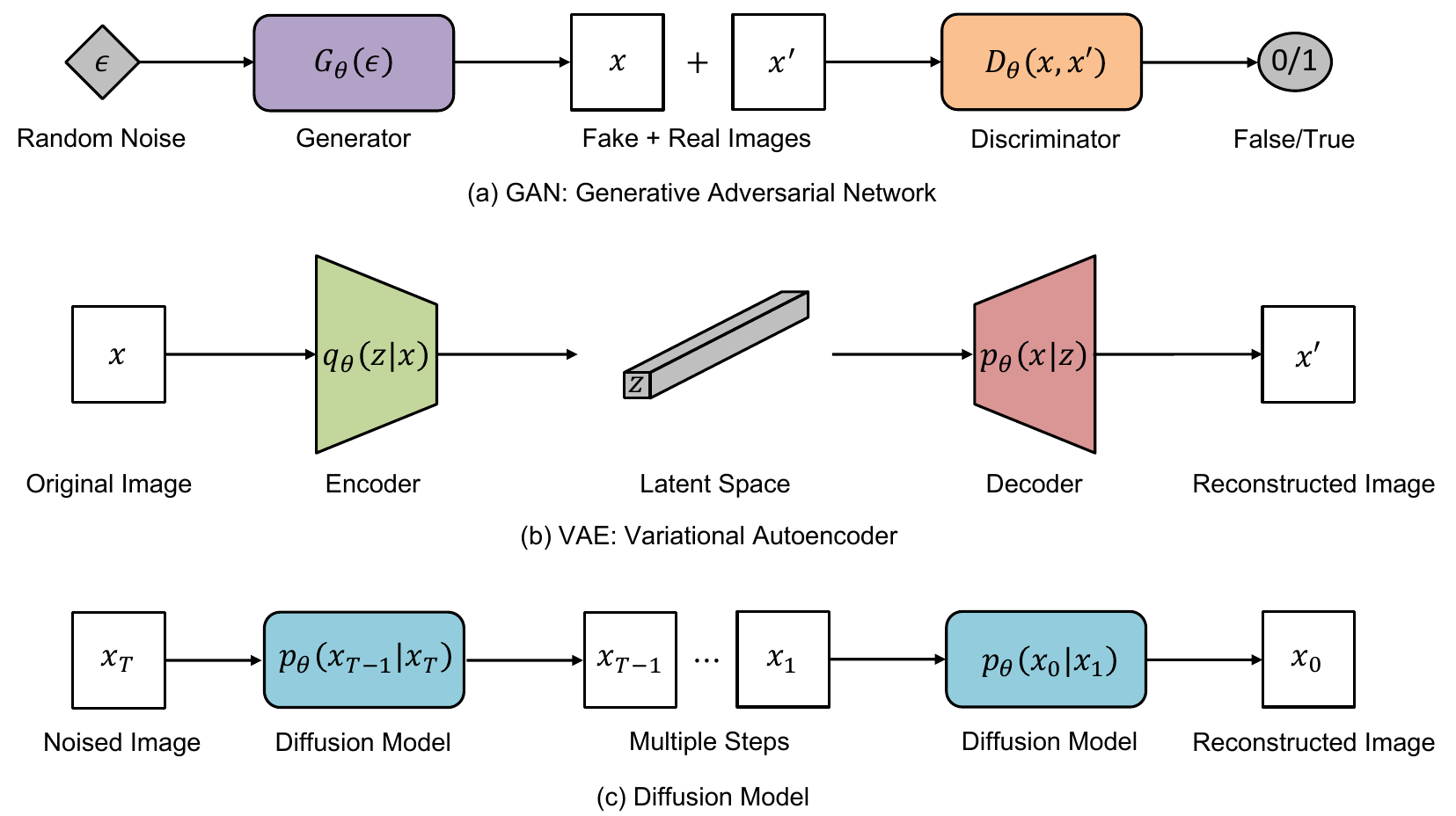}}
\caption{A sketch comparison of generative model architectures. 
(a) GAN model generates image using the generator that is trained by adversarial training. (b) VAE model reconstructs image by maximizing variational lower bound from the latent representation. (c) Diffusion model reconstructs image by denoising the Gaussian-noised image in an iterative process.\label{fig:generative-models}}
\end{figure*}
\subsubsection{GAN: Generative adversarial network}
GAN  \cite{GAN_Model} is a generative model that samples the image without the computation of the density function from the training data.
GAN consists of a \textit{generator} ($G$) and a \textit{discriminator} ($D$). The discriminator maximises the accuracy to distinguish the input images from the synthesised images with a binary output. The generator learns the image distribution from the training samples and tries to minimize the possibility to be classified as fake by the discriminator.

GAN training process is shown in Fig \ref{fig:generative-models}, which is based on the adversarial process with the minimax algorithm. This is achieved by either fixing the generator and taking gradient ascent on the discriminator:
\begin{equation}
   \nabla_{{\theta}_d} \frac{1}{n} \sum_{i=1}^n
   [logD_{\theta_{d}}(x^i)+log(1-D_{\theta_{d}}(G_{\theta_{g}}(z^i)))],
\end{equation}
or fixing the discriminator and taking gradient descent on the generator:
\begin{equation}
    \nabla_{{\theta}_g} \frac{1}{n} \sum_{i=1}^n
  log(1-D_{\theta_d}(G_{\theta_{g}}(z^i))),
\end{equation}
for a minibatch of $n$ samples of image data ${x_1,x_2,\dots,x_n}$ and noise data ${z_1,z_2,\dots,z_n}$.

Compared to VAE, GAN generates images with better fidelity. However, GAN could not extract an explicit representation of the data density $P(x)$,
and training a GAN model is trickier considering the slower converging trend. Moreover, there is also a need to balance the performance of discriminator and generator to avoid mode collapse that ends up generating the same sample from noises  \cite{Improved_Gan}. 
GAN model has achieved impressive results for image generation \cite{DCGAN,cycle_gan,StyleGan}, whose success also drives the development of text-to-image generation. Besides this, GAN model could also be applied to other complicated generation and perception tasks such as image super-resolution \cite{GAN_superresolution,BigGan_resolution,pyramid_gan_superresolution,coupled_gan}, and object detection  \cite{GAN_object_detection1,GAN_object_detection2}.

\subsubsection{Diffusion models}
Diffusion model \cite{DDPM,DDPM_Beats_Gan} is another type of generative model that could generate images with repeating steps, in a form of latent variables to model the probability distribution  \cite{Nonequilibrium_Thermodynamics_DPM}: $P(x_0) = \int dx(1...T)p(x(0...T)$, illustarted in Fig \ref{fig:generative-models}. Generally, a diffusion model involves adding noise to perturb the data during training and removing the noises with learned parameters during inference. Algorithm 1 and 2 summarize the general steps to train a diffusion model and apply a diffusion model to sample data, respectively. Using diffusion models for TTI not only ensures a high fidelity of the output image but also provides varieties of the image styles. Therefore, it has become one of the top choices for TTI generation. Apart from TTI generation, diffusion model also achieves impressive results in image segmentation  \cite{segdiff,label_segment,Implicit_segment,srdiff,diffuseq}, super-resolution  \cite{image_refinement,egsde}, natural language processing \cite{Diffusion_nlp,diffusionlm,latent_diffusion_text_model,diffsound}, or even in the field of reinforcement learning \cite{Diffusion_rl1,diffusion_rl2}. Recent researches have focused on improving diffusion model structures to increase the quality of the synthesised image or decrease the total time for the inference. There are two types of popular diffusion models, denoising diffusion probabilistic models (DDPM) and score-based diffusion models.


DDPM \cite{DDPM} is a fundamental type of diffusion model which uses discrete timesteps. DDPM consists of a forward and a reverse process, both of which follow the Markov chain process.
The forward process gradually adds noise to the original image until the data becomes a normal Gaussian noise. The noise added in each timestep is sampled from the normal Gaussian distribution and is scaled by a scheduler $\beta_t$ and $\alpha_t = 1-\beta_t$. This is normally done with an arbitrary \(T\) timesteps:
\begin{equation}\label{diff_forward}
q(x_{1:T}) = \prod_{t=1}^{T} q(x_t|x_{t-1}),\  q_{\theta} (x_{t} | x_{t-1}) = N(x_t,\sqrt{\alpha}_t, x_{t-1},\beta_t I).
\end{equation}
Because of the addition property of the Gaussian distribution, Eq. (\ref{diff_forward}) could be converted to:
\begin{equation}
q_{\theta} (x_{t} | x_{0}) = \mathcal{N}(x_t,\sqrt{\Bar{\alpha_t}} x_0,(1-\Bar{\alpha_t}) I)\label{eq:ddpm-forward},
\end{equation}
where $\Bar{\alpha_t} = \prod_{s=1}^{t}\alpha_s$. This ensures that we can directly calculate data at any timesteps with a noise sampled from the normal Gaussian distribution: $x_t(x_0,\epsilon)=\sqrt{\Bar{\alpha_t}} x_0 + \epsilon \times \sqrt{(1-\Bar{\alpha_t})}$. The scheduler design of $\alpha$ ensures that  $\Bar{\alpha_n} = 0 $, which indicates that the data will be perturbed to a normal Guassian noise $\mathcal{N}(0,I)$ at the last timestep $n$.

The reverse process is also a Markovian chain that samples the data from a learned Gaussian distribution at each timestep to remove the noise:
\begin{equation}
p_{\theta} (x_{t-1} | x_{t}) = \mathcal{N}(x_t-1,\mu_\theta(x_t,t),\Sigma_\theta(x_t,t)).
\end{equation}

\begin{algorithm}[t]
	\caption{Diffusion model training} 
	\begin{algorithmic}[1]
		\For {every training iteration}
            \State Sample $t$ from discrete timestep \cite{DDPM} \cite{DDPM_Beats_Gan}, \emph{i.e.,} $1,2,\ldots ,T $, or from continuous timestep \cite{DDIM} \cite{Score_based_DDPM} \emph{i.e.,} $t\sim[0,1]$.
            \State Sample random noise from $\epsilon \sim \mathcal{N}(0,1)$.
			\State Calculate $x_t$ based on DDPM forward Eq. (\ref{eq:ddpm-forward}) or SDE forward Eq. (\ref{eq:sde-forward}).
			\State Update the model with noise prediction $\epsilon(x_t,t)$ or score function $s(x_t,t)$.
		\EndFor
	\end{algorithmic} 
\end{algorithm}

DDPM takes the variance as a fixed schedule guided by $\beta$:
$\Sigma_\theta(x_t,t) = \beta_t$ or $\Sigma_\theta(x_t,t) = \frac{1-\Bar{\alpha_t}}{1-\Bar{\alpha_t-1}}\beta_t$ and calculates the mean value with the approximation of $x_0$ given $x_t$:
\begin{equation}\label{eq10}
\mu_\theta(x_t,t) = \Bar{\mu}(x_t,\Bar{x_0}) = \frac{1}{\sqrt{\alpha_t}}
(x_t-\frac{\beta}{\sqrt{1-\Bar{\alpha_t}}}\epsilon_t(x_t,t)),
\end{equation}
where Eq. (\ref{eq10}) is derived from the Bayesian theorem with $\epsilon_t(x_t,t)$ predicting the Gaussian noise added in the forward process at the timestep $t$. Original image $x_0$ could be approximated by $\boldsymbol{x}_0 = \frac{1}{\bar{\alpha}_t}\left(\boldsymbol{x}_t  - \bar{\beta}_t \boldsymbol{\epsilon_t}\right)$ at any timestep.

Training is based on the optimisation of the variational upper bound, which could be reparameterized based on the form of Eq. (\ref{eq10}):
\begin{equation}\label{diff_loss}
L = \left\Vert\boldsymbol{\epsilon} - \epsilon_{\theta}(\sqrt{\Bar{\alpha_t}} x_0 + \epsilon \times \sqrt{(1-\Bar{\alpha_t})}, t)\right\Vert^2.
\end{equation}
The loss function's purpose is to calculate the MSE error between the predicted noise at a certain timestep and the noise that is sampled from the forward process.

DDPM uses a time-aware U-net \cite{U-net} as the backbone network that takes the noised data $x_t$ and the timestep \(t\) as the input. The output is the predicted noise in the same dimension as the input, corresponding to Eq. (\ref{eq10}) and (\ref{diff_loss}). This backbone architecture is adopted by a number of upcoming diffusion models \cite{Score_based_DDPM,DDIM,Analytic-Dpm} and diffusion-based TTI generation models \cite{Dalle2,Imagen,stable_diffusion}.

DDPM shows a much more stable training process in comparison to GAN, and keeps high fidelity of the image at the same time. However, sampling noise at the first stage of the diffusion process often performs poorly on approximating $x_0$, therefore, it usually needs massive repeating steps to adjust the sampled noise and thus denoise the data towards the correct direction.

\begin{algorithm}[t]
	\caption{Diffusion model Inference} 
	\begin{algorithmic}[1]
		\State Sample $x_t$ from normal gaussian distribution $x_t \sim \mathcal{N}(0,I)$.
            \State Sample discrete timesteps from $1,2,\ldots,T$ or continuous timestep from [0,1].
            \For{t in Reverse(timesteps)}
                \State Calculate the noise distribution \cite{DDPM} \cite{DDPM_Beats_Gan} \cite{DDIM} $\epsilon(x_t,t)$ or score function \cite{Score_based_DDPM} $s(x_t,t)$ with the corresponding diffusion model.
                \State Approximate $x_{t-1}$ or $x_{t-\Delta t}$ based on the reverse function.
            \EndFor
	\end{algorithmic} 
\end{algorithm}

Score-based diffusion models \cite{Score_based_DDPM} generalize the DDPM process using differential equations. Along a similar vein, score matching \cite{Score_based_generative_process} refers to the use of a score function to estimate the gradient of the dense probability function $S(x_t,t) = \nabla_x logP(x_t)$. In diffusion models, this could be applied using stochastic differential equations (SDE) \cite{Score_based_DDPM}:
\begin{equation}
dx = f_t(x) dt + g_t dw\label{eq:sde-forward},
\end{equation}
for the forward process. There are a few ways to estimate the score through the training of a score-based model, such as denoising score matching \cite{Score_based_generative_process,denoising_score_matching}, sliced score-matching \cite{sliced_score_matching}, and finite-difference score matching \cite{Finite_difference_score_matching}. SDE diffusion model \cite{Score_based_DDPM} takes the denoising score matching approach via continuous time generalization of the DDPM loss function to train the score model:
\begin{equation}
\begin{split}
    \theta^\star = {\rm argmin}_\theta\ \mathcal{E}_t\lambda(t)\mathcal{E}_{x(0)}\mathcal{E}_{x(t)|x(0)}[||s_\theta(x(t),t)\\-\nabla_{x(t)}\log P(x_t|x_0)||^2].
\end{split}
\end{equation}

The reverse process also matches a diffusion process \cite{Reverse_diffusion}, yielding the result of:
\begin{equation}dx = \left[f_t(x) - g_t^2\nabla_{\boldsymbol{x}}\log p_t(x) \right] dt + g_t dw\label{eq:sde-reverse}.\end{equation}
Such reverse process could alternatively be represented by an ordinary differential equation (ODE):
\begin{equation}
d{x} = \left({f}({x,t}) - \frac{1}{2}g_t^2\nabla_{x}\log p_t(x)\right) dt,
\end{equation}
with all of them based on the trained neural network predicting dense probability function to denoise the image towards the correct direction.

\subsection{Large Language Model}
\begin{table*}[h]
\begin{center}
\caption{A comparison of the popular large language models and vision models across literature. As you can see there is a trend for much large parameters, however, as the computational cost grows there is limited compute power to meet such a large demand. \label{Large language table}}
\begin{tabular}{l|ccc}
\toprule
model & Size & Model Type \\
\midrule
CLIP \cite{Clip} &  151M &  Language-vision contrastive model\\
BERT \cite{Bert} & 340M &  Encoder-only based Transformer model architecture\\
Megatron-LM\cite{megatronlm} & 8.3B  &  Encoder-decoder based large Transformer model\\
T5 \cite{T5} & 11B  &  Encoder-decoder based Transformer model architecture\\
Flan-T5 \cite{FlanT5}& 11B & Improved T5 model\\
OPT \cite{opt} & 125M-175B & Decoder-only pre-trained transformers with varied model parameters\\
LLama-1,2 \cite{touvron2023llama1, touvron2023llama2} & 3B-70B & Large autoregressive model with designed MLP architecture.\\
GPT-3 \cite{GPT-3} & 175B & Large autoregressive model with GPT structure.\\
Megatron-Turing \cite{MT-NLG} & 530B & Large autoregressive model with GPT structure.\\
GPT-4 \cite{gpt4} & - & - \\
\bottomrule
\end{tabular}
\end{center}
\end{table*}
Language models are advanced natural language processing (NLP) systems that are capable of processing and generating human-like languages. Through the use of deep learning, it has allowed for language models to learn directly from raw text data, without the need for extensive feature engineering. The breakthrough of language model comes with the introduction of the Transformer \cite{Transformer} with an attention mechanism that remembers and builds the relationship based on a complete text sequence. This is instrumental for the emergence of large language models that are trained on massive amounts of data to capture the nuances and complexities of human language. With architectural improvements and significant scaling in model size, large language models have brought impressive performance in a wide range of language tasks such as machine translation, question-answering, and dialogue system. There are a few variations in the latest LLMs, such as BERT-based encoder-only language models \cite{Bert,albert,distilbert,UniLM,electra,ernie,ernie-titan,deberta}, encoder-decoder language models \cite{T5,T0,glm130b,alexatm,TKinstruct,xue2021mt5,mT0,stmoe}, and GPT-based language models \cite{GPT-1,GPT-2,GPT-3,zhang2022opt,chowdhery2022palm,bloom,lamda,LLaMa,gpt-neoX,bloomberggpt,Gopher,MT-NLG,Chinchila,webgpt}. These variations are succinctly presented in Table \ref{Large language table}, which enumerates a subset of widely recognized large language models. Notably, there is a discernible trend of exponential growth in model size across these models.

LLMs have emerged as a prominent approach in natural language processing, wherein the size of small language models is scaled up significantly. For instance, GPT-3  \cite{GPT-1} substantially increases the parameters of the GPT structure from 117M to a staggering 175B. The training of LLMs heavily relies on parallel strategies and distributed computing, with several frameworks capable of supporting large-scale training, including DeepSpeed  \cite{deepspeedmoe,Deepspeedzero}, Megatron  \cite{megatronlm}, and colossal-ai  \cite{colossalai}. The expansion in model size enables LLMs to possess capabilities that are absent in smaller language models  \cite{LLMSurvey}. One notable advantage of LLMs is their ability to perform in-context training, which allows the model to process examples before engaging in a specific task. Additionally, LLMs have demonstrated proficiency in solving intricate reasoning problems, such as mathematical induction \cite{naturalprover}.

Text-to-image models rely on text encoders to convert textual sentences into meaningful representations that can be utilized for image generation. An effective text encoder plays a crucial role in capturing the essential semantic and contextual information contained within the text description. This information is then utilized to guide the generation process of the output image. Recent advancements in LLMs have introduced a diverse array of pre-trained text encoders that can be seamlessly integrated into generative models for image generation tasks \cite{DALLE,Imagen,Parti,stable_diffusion,muse}. These models have already demonstrated remarkable capabilities in language feature engineering, which align well with the requirements of TTI generation. Leveraging these pre-trained encoders not only saves substantial time during TTI model's training, but also eliminates the need to invest excessive efforts in improving the model's performance for text embedding.

\subsection{Large Vision Model}
The advancements of language model architecture also spur their application to the research of vision models \cite{ViT,localvit,Visionmoe,vlmo,DeiT,TiT_ViT,pyramid_Vit} that are able to learn pixel-level information by encoding image patches with a Transformer-based encoder receiving spatial inputs rather than textual tokens. Vision models could be used in varied visual tasks such as image classification \cite{ViT}, object detection \cite{Detr,Yolos}, and segmentation \cite{panoptic,CMSA,SETR}. The work of language models and vision models could be further extended to multi-modal learning \cite{beit,Clip,imagebert,unifying_VL,xgpt,Declip,vilt,pixelbert,visualbert,visualgpt,interbert,vlmixer,kaleidobert} that finetunes image-text data with the combination of language and vision models to understand the connections among different types of data. Thus, multi-modal learning could generate modalities different from the input data, such as images to captions.

Vision models could also be used as auto-regression models \cite{Dalle2,Parti} that encode images to a latent space representation for contrastive learning with text embeddings.  Moreover, the integration of language and vision models has propelled the adoption of multi-modal diffusion for TTI tasks. These models effectively connect information from text embeddings and image embeddings, facilitating the diffusion process not only from text embeddings to image embeddings but also from image embeddings to ground truth image output.

Among all the language and vision models, CLIP \cite{Clip} stands out as the preeminent choice, assuming a pivotal role in the construction of TTI generation. Originally conceived as a language-vision model, CLIP was initially tailored for tasks such as image classification, action recognition, and optical character recognition. CLIP \cite{Clip} consists of a Transformer language encoder \cite{Transformer} and a ResNet or Vision-Transformer image encoder \cite{Improved-Resnet,ViT}. Both aforementioned models map the input data to the latent space sharing the same embedding dimension. Training is based on the objective of maximizing the dot-product of text embeddings and image embeddings with the correct image-text pair, and minimizing the value of the incorrect ones, as indicated by Algorithm 1. In this way, CLIP shows impressive ability in matching images with any text inputs based on the highest dot-product value. CLIP also provides strong zero-shot learning abilities when transferring the model to downstream tasks. It is able to produce high accuracy classification results in zero-shot image datasets. 
\begin{algorithm}
	\caption{CLIP Model}
    \label{alg:algo1}
	\begin{algorithmic}[1]
		\For {one step during training}
				\State Sample a batch of (Image,Text) pairs with size N from the dataset.
				\State Feed images through the image encoder to get image embeddings and map them to multimodal embeddings $I_1,I_2,\ldots,I_N$
			\State Feed texts through the text encoder to get text embeddings and map them to multimodal embeddings $T_1,T_2,\ldots,T_N$
			\State Create the cosine similarity matrix with dimension $N\times N$.
            \State Optimize the model based on the cross entropy that maximizes similarity scores over $N$ matched image-text pairs and minimizes similarity scores over $N^2-N$ unmatched pairs. 
		\EndFor
	\end{algorithmic} 
\end{algorithm}
CLIP has become a cornerstone for the development of TTI generation. A pretrained CLIP text-encoder provides strong capability for text encoding and zero-shot representation while keeping a resonably small model size, therefore, it is widely used for Diffusion TTI models \cite{stable_diffusion,Dalle2,Corgi,ControlDiffusion} and recent GAN TTI models \cite{galip,Gan_with_Clip,GigaGan}. Besides, CLIP enables multi-modal learning for contrastive text-image representations, which has spurred the development of prior-decoder TTI architecture \cite{Dalle2,Corgi,spatext} that generates images without arbitrary need of text embeddings. 
\section{Text-to-Image generation}

\begin{figure}[H]
\centering 
\includegraphics[scale=0.43]{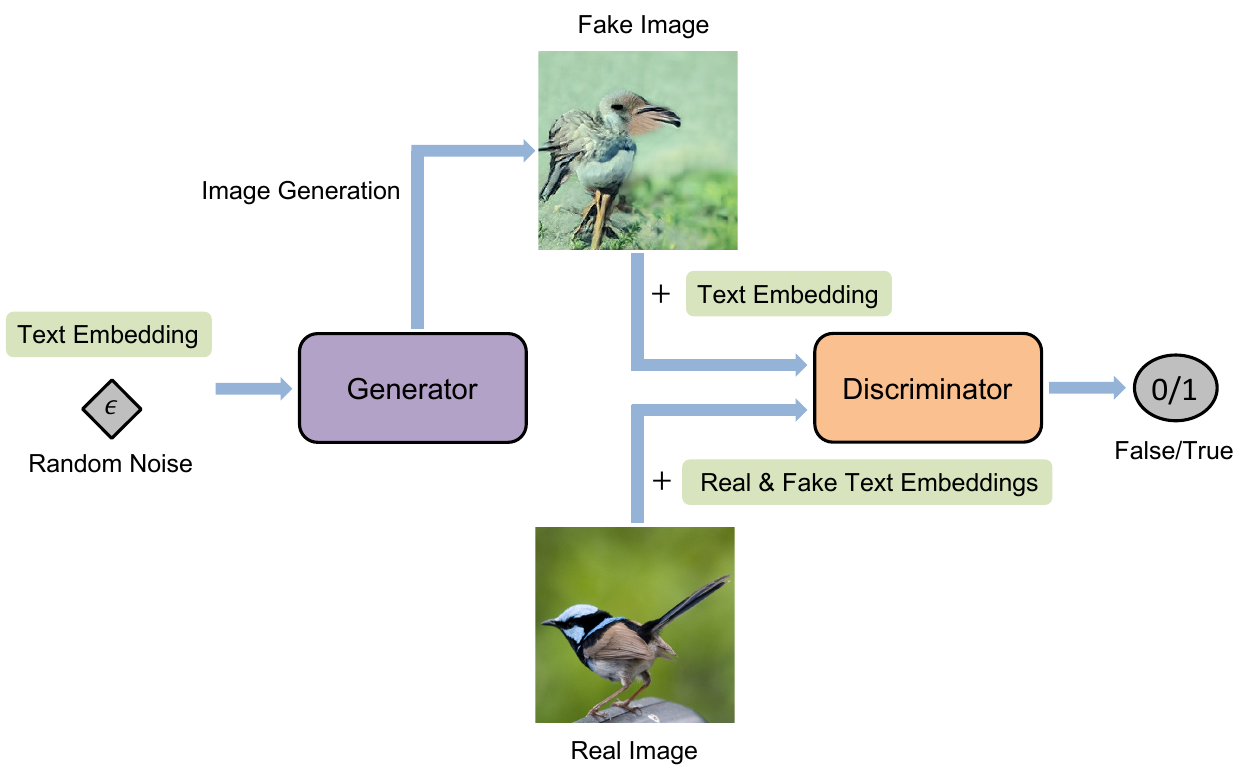} 
\caption{A general structure of GAN-based text-to-image generation models, where the Generator takes text description as input and tries to generate fake images that 'fool' the Discriminator, and the Discriminator learns to discriminate images using fake images with real text, and real images with real and fake texts.} 
\label{Fig.ganTTI}
\end{figure}

\subsection{Prior to Large Model Era}
Text-to-image generation could be traced back to the emergence of GAN models \cite{Gan-CLS,stackgan,AttentionGAN,original_gan_tti,GAWWN}, and the performance is further improved by architecture and algorithm updates in the field \cite{XMC-GAN,TreCS,DCGAN,MirrorGan,SEGAN,HDGAN,PPAN,SDGAN,li2022eliminating}.
The development of GAN-based TTI models has usually involved in generating images with relatively fewer model parameters, and is limited to a domain specific set of dataset \cite{Cifar10,COCO,Imagenet}. Early models have already shown the possibility to generate images based on the text description. However, there exists a huge gap of image quality and diversity compared to the large TTI models. Due to the limitation of the size of the dataset and the performance of text encoders, GAN-based TTI is usually not capable of generating complicated and detailed images which are more likely to be sought after by user descriptions. Furthermore, GAN models do not have the capability to capture zero-shot text descriptions as well as diffusion TTI models. Despite the drawbacks for early TTI models, consistent research has improved the algorithms and architectures, which not only makes text-to-image generation possible, but enhances the image quality to a competitive level against state-of-the-art TTI models. In this section, we will list examples of GAN-based TTI models and elucidate the features that support text-to-image generation where some of them still impact on the creation of new models in the field.

\subsubsection{Image refinement}
Applying GAN models for text-to-image generation can be straightforwardly achieved through the incorporation of additional conditional information \cite{conditional_gan}. This process is typically implemented by taking text embeddings with Gaussian noise as inputs to the generator, images and text embeddings to the discriminator, as depicted in Fig \ref{Fig.ganTTI}. This ensures that the GAN model can possess the generative ability to align its output with the text descriptions, instead of creating random images from Gaussian noise unrelated to the input text. While this has been widely adopted in some prominent work \cite{original_gan_tti}, many new techniques are brought up subsequently that aim at enhancing both model performance and the quality of generated images. One common way is the stacking of multiple GAN components for image refinement. For example, StackGAN \cite{stackgan} is composed of two stages of GAN models. The first-stage generator takes the text embeddings from an LSTM encoder \cite{LSTM} and produces the images with $64\times64$ resolution. These smaller images are then fed into the first-stage discriminator with real image and text embeddings. The second-stage GAN repeats the same process whereas it generates the images with a better fidelity of $256\times256$ resolution. 

StackGAN++ \cite{stackgan++} improves the structure of StackGAN with a three-stage tree-like hierarchy. At the start of each stage, there are concat, residual and upsampling layers to pre-process inputs and generate image features for the next generator. Each of the three generators takes the up-sampled image features and produces images with size \(64\), \(128\) and \(256\), respectively. Besides, both StackGan and StackGan++ implement conditional augmentation that adds random Gaussian noise to the text embeddings as new conditional variables. This technique ensures that the model is able to have a stronger noise-proof ability and generate a wider variety of images.

\subsubsection{Image-text alignments}
Achieving precise alignment between text and image poses a significant challenge in TTI generation, a task that is more complex compared to other image generation applications. AttnGAN \cite{AttentionGAN} proposes an effective way to measure the relevance of image-text pairs. As a similar approach to StackGAN++, AttnGAN \cite{AttentionGAN} takes three refinement stages with conditional augmentation and pre-processing layers, and samples the image from size \(64\) to \(256\). Two new features are introduced for image-text alignment. One is the introduction of the attention model (different from the attention mechanism) and the other one is a pre-trained Deep Attentional Multimodal Similarity Model (DAMSM) which takes in additional word features procured by the AttnGAN.  In the DAMSM model, these extra word features contribute to quantifying the similarity between textual descriptions and synthesized images. This is done by calculating the relevance between words and each feature vector extracted from a pre-trained image encoder \cite{Imagenet}. During training, AttnGAN extracts these additional word features together with text embeddings from a bi-directional LSTM \cite{LSTM}. Text embeddings are taken as conditional information and concatenated with noise as model input, and the word features are used within both DAMSM and attention models. In the attention model, word features are used to compute the word-context vector for each image feature representing the relevance of an image sub-region to each word, and then word-context vectors are fed into the next stage together with image features. 

AttnGAN uses matching score $R(Q,D)$ for the image-text similarity calculated from the DASAM model, which could then derive the DASAM loss function:
\begin{equation}
\begin{split}
    L_{DASAM} = -\sum_{i=1}^M \log P(D_i|Q_i)-\sum_{i=1}^M \log P(Q_i|D_i) \\-\sum_{i=1}^M \log P(\hat{D_i}|\hat{Q_i})
    -\sum_{i=1}^M \log P(\hat{Q_i}|\hat{D_i}),
\end{split}
\end{equation}
where $P(D_i|Q_i) = \frac{\exp(\gamma R(Q_i,D_i))}{\sum_{j=1}^m \exp(\gamma R(Q_i,D_j))}$ refers to the probability that image $Q_i$ is
matched with its corresponding text descriptions $D_i$. They are calculated in a level of pixel and word embeddings. $P(\hat{D_i}|\hat{Q_i})$ refers to the probability considered in the global image and sentence level. A DASAM model absorbs more information to compare the similarity between texts and images. Therefore, adding a DASAM loss function upon training becomes a stronger way to align the images with text descriptions.

Cross-modal contrastive generative adversarial network (XMC-GAN) \cite{XMC-GAN} is another model that considers regional features and attempts to enhance semantic fidelity with text descriptions. Unlike AttnGAN, XMC-GAN introduces an attentional self-modulation layer in the generator that takes use of both word embeddings and sentence embeddings as conditional inputs to generate finer-grained image regions. Besides, the discriminator also extracts features of both regional and global features from synthetic and real images to compare the mutual information with text and sentence embeddings. In addition to the general loss function of GAN model, XMC-GAN introduces contrastive losses that could maximize the lower bound of mutual information of: sentence embeddings and both real and fake images, word embeddings and both real and fake images' regions, and also the mutual information of real and fake images. 
\begin{figure*}[htp] \centering
\includegraphics[scale=0.65]{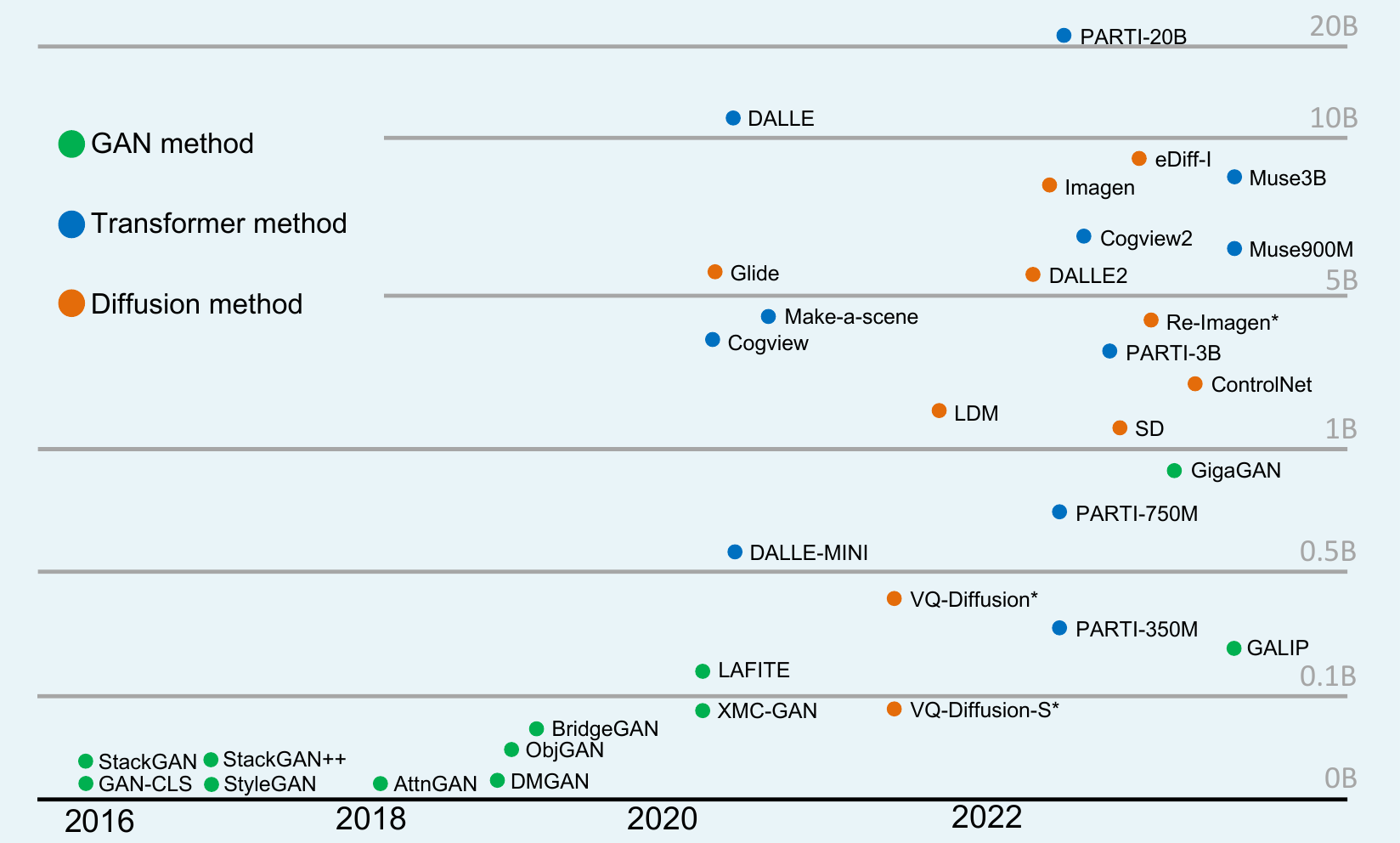}
\caption{Timeline of TTI model development, where green dots are GAN TTI models, blue dots are autoregressive Transformers and orange dots are Diffusion TTI models. Models are separated by their parameter, which are in general counted for all their components. Models with asterisk are calculated without the involvement of their text encoders.}
\label{Fig.timeline}
\end{figure*}
\subsubsection{Image diversity enhancement}
Image diversity is always a vital evaluation criterion of TTI models as text-image translation is not rigorously a one-to-one mapping relationship. For example, one image could be represented by a different form of text expressions and the same semantic text descriptions could in theory represent different results in the output space. Early TTI models have the ability to generate images that align with text descriptions but discriminate images in diverse forms. Several research has targeted on and solve such problems \cite{AC-GAN,TAC-GAN,Scene_graph_Gan,MirrorGan}. AC-GAN \cite{AC-GAN} modifies the general GAN structure to produce label probability distribution inside the generator together with the True/Fake predictions. Such improvements have allowed synthetic images to have an overall diversity coming from all labelled classes. This architecture is also adopted for text-to-image synthesis by TAC-GAN \cite{TAC-GAN} with additional text inputs to both generator and discriminator. The discriminator components keep the layer to predict the probability distribution of class labels receiving fake images and text. This results in a better image diversity compared to the earlier GAN-based TTI generation models using only text embeddings such as GAN-CLS \cite{Gan-CLS} and StackGAN \cite{stackgan}. \\

\subsection{Large Model with GAN Method}
Recent advancements in large generative models have brought many opportunities to improve the performance of GAN-based TTI models, often accompanied by an escalation in model size, as depicted in Figure \ref{Fig.timeline}. Notably, one primary avenue of advancement lies in the adoption of large language models. The evolution of these robust linguistic architectures has imbued GAN models with the capacity to deploy more potent text encoders. This shift signifies a departure from conventional LSTM-based text encoders, frequently employed within previous GAN frameworks. For example, there has been researches that integrate GAN models with the pretrained CLIP model as a powerful tool for text-encoding, visual feature extraction and image-text alignment \cite{Gan_with_Clip,lafite,stylegannada,fusedream,galip}. CLIPGLaSS \cite{Gan_with_Clip} uses a genetic algorithm to ensure that images generated from Generator has the maximized CLIP score with their corresponding text prompts. Similarly, FuseDream \cite{fusedream} aims at maximizing an augmented CLIP score that is more robust against adversial attacks. Both LAFITE \cite{lafite} and GALIP \cite{galip} uses CLIP model to adjust loss function for generator or discriminator to ensure GAN feature space is aligned with the pretrained CLIP. 

Furthermore, augmenting model size and expanding training datasets have emerged as additional avenues for performance enhancement. GigaGAN \cite{GigaGan}, for instance, achieved this by scaling up model parameters by approximately 20-fold compared to its foundational counterpart \cite{stylegan2}. Its training process leveraged expansive datasets such as LAION-2B \cite{LAION-5B} and COYO-700M \cite{coyo-700m}. Collectively, these enhancements usher in noteworthy progress for GAN TTI models, rendering their generative outcomes on par with the latest state-of-the-art autoregressive Transformer and diffusion-based TTI models.

\subsection{Large Model with Autoregressive  Method}
Transformer models have been applied broadly in both natural language processing and computer vision, and have shown an impressive result in text-to-image generation that gains the performance as close as the SOTA diffusion models. Autoregressive TTI model refers to the use of an autoregressive Transformer to predict image tokens from text. This is always employed with another encoder-decoder model that could encode an image to tokens and decode back to the image, and this structure plays a vital role in training the autoregressive Transformer. In recent years, with the development of Transformer-based models such as Vision Transformers and GPT, applying autoregression models to TTI produces impressive results. As shown in Fig \ref{Fig.timeline}, large autoregressive Transformer has become one of the dominant models in recent years.
\subsubsection{DALLE}
DALLE \cite{DALLE} is one of the latest text-to-image models that achieves impressive results with large models and datasets. The first stage involves training a discrete variational-autoencoder (dVAE) \cite{discrete_vae} that maps an image into $32\times32$ tokens where each grid has \(8192\) possible values. This is able to compress the $256\times256$ image data into a much smaller dimension with only slight information loss. In the next stage, text is encoded with byte pair encoding (BPE) \cite{BPE-Encoding} to tokens of size \(256\), and concatenated with image tokens. The new tokens are then fed into a decoder-only autoregressive Transformer to compute text-image joint distribution. This Transformer consists of \(64\) self-attention layers with casual mask for text attention and either row, column or convolution mask for spatial image attention.

Training of the overall procedure is based on maximizing the evidence lower bound \cite{Auto-Encoding,VAE_Inference} of the text-image distribution:
\begin{equation}
\begin{split}
    \log p_{\theta,\psi}(x,y) \geq  \mathop{\mathbb{E}}_z(\log p_\theta(x|y,z)\\ - \beta D_{KL}(q_\phi(y,z|x)||p_\psi(y,z))).
\end{split}
\end{equation}
The attainment of this goal unfolds in a two-stage process. Primarily, the dVAE undergoes training to optimize the parameters $\theta$ and $\phi$. Subsequently, once these parameters are fixed, the autoregressive Transformer is engaged in training to maximize the aforementioned lower bound with respect to parameter $\psi$. To optimise the probability distribution of discrete parameters, they use the gumble-softmax funciton with an annealing factor $\tau = \frac{1}{16}$ to calculate a relaxed ELB when training the dVAE. When it comes to the training of the autoregressive Transformer, the image tokens will be sampled from the dVAE directly via an \verb|argmax| operation without any gumble noises.

There are a few generics models of DALLE. DALLE-mini \cite{dayma_2021_dalle_mini} has only 4B parameters which are 27 times less than the original model. It has similar structures as DALLE whereas using BART models as the text encoder and autoregressive image token decoder. Besides, DALLE-mini replaces dVAE with a VQGAN \cite{VQ_GAN} for image encoding and decoding. Training of the BART decoder is based on the cross-entropy loss between the image tokens generated by BART decoder and VQGAN encoder. CLIP model is applied for choosing the best one among the images sampled from the VQGAN decoder. In the later stage, there is a successful development of DALLE-mega which is an enlarged version of DALLE-mini.

\subsubsection{Cogview}
Cogview \cite{Cogview} improves upon the work of DALLE with a sophisticated set of features to address DALLE drawbacks. Cogview resembles the structure of DALLE, however, the backbone is an undirectional Transformer(GPT) and the image reconstruction model is a discrete autoencoder. Besides, compared to DALLE, Cogview develops a more efficient and general approach to stabilize the training process by regularizing the values of the LayerNorm and Attention operations of the Transformer.

Apart from the above innovation, Cogview \cite{Cogview} finetunes the model to adapt to more text-image tasks such as super-resolution, image captioning, and style changing, and achieves impressive results. The finetuning here allows the TTI model to extend its understanding of texts and images to a wide range of similar tasks to show a good ability of generalisation in the computer vision field. This philosophy is widely accepted by future text-to-image models.       
\subsubsection{PARTI}
Google PARTI \cite{Parti} is also a recent TTI model that can synthesize images with high fidelity, varied styles and compositions. Compared to its previous Imagen model, PARTI performs better in controlling the generation direction towards the target image with longer and more detailed text description. In addition to this, it achieves consistent image improvements by scaling up its Transformer models. PARTI has greatly boosted the image quality with the zero-shot FID score of 7.23 and finetuned FID score of 3.22 on MS-COCO. This has again proved the feasibility of using autoregressive models to achieve top performance in TTI generation.

PARTI also works in a very similar way to DALLE. It uses a vision Transformer \cite{ViT} based model (VIT-VQGAN) as image tokenizer (encoder) and detokenizer (decoder). Alongside the encoder and decoder, there is also an autoregressive Transformer to encode text prompts and transfer them to image tokens. However, this Transformer is different from DALLE in that PARTI trains its own text-encoder in order to achieve a stronger textual representation. The first training stage involves training of the VIT-VQGAN \cite{Improved_VQGAN} tokenizers to encode an image of size $256\times256$ to discrete visual tokens with a fixed length of 1024, and the detokenizer's purpose is to reconstruct the image tokens 
back to the image. PARTI adopts the VIT-VQGAN structure \cite{Improved_VQGAN} and starts with a smaller configuration. Later on the tokenizer will be frozen and the detokenizer will be replaced with a larger sized and fine-tuned version for better visual acuity. In addition to this, a super-resolution model with 30M parameters is stacked onto the VIT-VQGAN model to convert the $256\times256$ image outputs to the size of $1024\times1024$ for a better visual illustration.

After the completion of training the tokenizer, the image tokens are then used to train the autoregressive Transformer. PARTI takes text-token inputs with a max length of \(128\) from a sentence-piece model with a vocubulary size of \(16,000\) sampled from the training data's corpus. This model implements the masked sparse attention mechanism for both encoder and decoder of the autoregressive model. Training is performed on the models with parameters ranging from \(350M\) to \(20B\), which is done by scaling up the MLP dimensions with an expansion ratio of \(4\) and also doubling the attention heads whenever the MLP dimensions are doubled. This scaling of parameters within training has helped to explore TTI results on a wide range of model sizes.
\subsubsection{Further Improvements to Autoregressive Architecture}
Despite the notable achievements of recent autoregressive Transformers, their image token generation still relies on hundreds of iterative steps. To overcome this limitation, a potential enhancement involves exploring the utilization of non-autoregressive Transformers to achieve increased decoding parallelism. An example of such an approach is showcased in the work of MUSE \cite{muse}. MUSE \cite{muse} employs masked transformers that apply partial masking to image tokens during training. This methodology predicts the marginal distribution of masked tokens while leveraging unmasked text embeddings as conditional inputs. This training process encompasses both their base text-to-image Transformer and the super-resolution Transformer.

During inference, the model follows an iterative parallel decoding approach to predict all masked image tokens at each step. To improve image quality, only tokens with the highest prediction confidence are unmasked and used as inputs for subsequent steps. In contrast to autoregressive Transformers, which necessitate at least 256 steps, MUSE requires only 24 decoding steps for the base Transformer and 8 steps for the super-resolution Transformer. This design not only alleviates the time efficiency concerns of autoregressive Transformers through enhanced parallelism but also yields impressive text-to-image outcomes. For the purpose of comparative analysis in subsequent sections, MUSE is grouped with autoregressive Transformers.
\begin{figure*}[htp] \centering{
\includegraphics[scale=0.6]{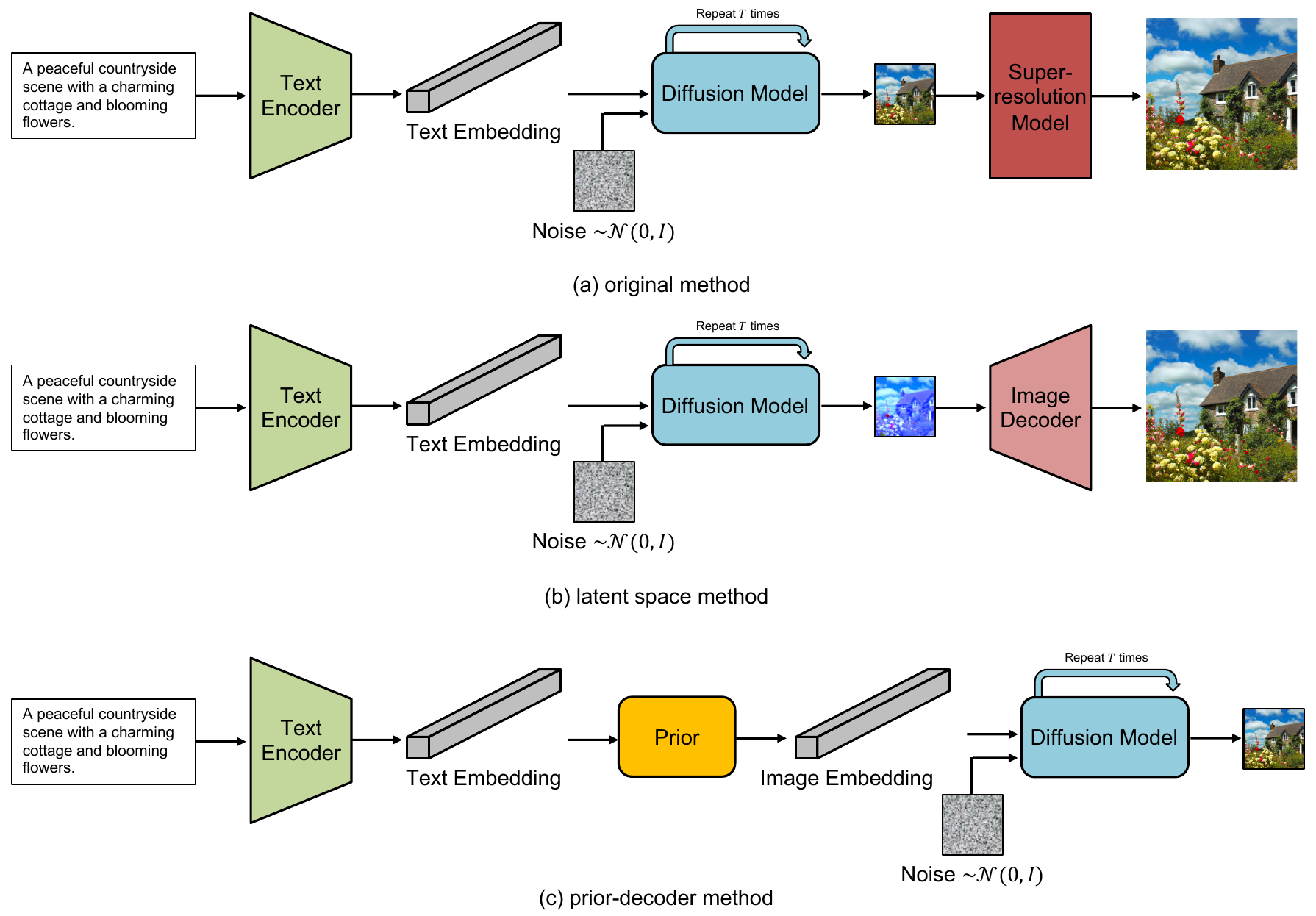}}
\caption{Diffusion TTI model architectures, with original diffusion structure shown in (a), latent space structure shown in (b) and prior-decoder structure shown in (c), note that in prior-decoder structure, the output image has relatively small scale, therefore superresolution model could be also applied to this structure. }
\label{fig:diffusiontti}
\end{figure*}
\subsection{Large Model with Diffusion Method}
Diffusion models for TTI generation involve combing an extra conditional element to the U-net. This is usually achieved by encoding text descriptions with a strong text encoder, such as CLIP, Google T5 \cite{Clip,T5}, etc. Besides, \cite{DDPM_Beats_Gan} introduces classifier-guided diffusion process, which guides the diffusion process towards the direction of the class label as shown in the following equation:
\begin{equation}\label{eq:classifier-guidance}
    \hat{\mu}(x_t|y) = \mu_{\theta}(x_t|y) + s*\Sigma_{\theta}(x_t|y)\nabla_{x_t}\log p_\phi(y|x_t),
\end{equation}
where the mean value of predicted image distribution is further perturbed by the gradient of log probability of a target class predicted by a classifier, $s$ is called guidance scale. 

Classifier guided diffusion models have been further improved by  \cite{Classifier-free_diffusion} that introduces classifier-free guidance, in which a diffusion model is jointly trained with and without conditional information whilst changing the noise based on a scale factor. Showcased in Eq. (\ref{eq:classifier-free-guidance}), the inference could rely solely on the diffusion process without the need for classifier models:
\begin{equation}\label{eq:classifier-free-guidance}
    \hat{\epsilon}(x_t|y) = \epsilon_{\theta}(x_t|\emptyset) + s*(\epsilon_\theta(x_t|y)-\epsilon_{\theta}(x_t|\emptyset)).
\end{equation}

Together with the above methods, diffusion models pose impressive results in sampling images based on the text description. Figure~\ref{fig:diffusiontti} illustrated the most common diffusion TTI architectures. In this section we will explain in detail about different TTI generation models that take use of the diffusion method.
\subsubsection{Early TTI applications with diffusion}
GLIDE \cite{Glide} is one of the first applications of diffusion models for TTI generation. It applies a fairly simple approach by generating a random Gaussian noise and inserting it into the diffusion model together with a CLIP-encoded text variable. Both classifier-guided and classifier-free diffusion are implemented during the training process for comparison. GLIDE also adopts the CLIP model \cite{Clip} to the guided diffusion function. 
The guidance function is provided for calculating the similarity between the noise image and text caption in a certain timestamp. The process of the guidance function can be detailed by: 
\begin{equation}\label{eq:glide-process}
    \hat{\mu}(x_t|c) = \mu_{\theta(x_t|c)} + s*\Sigma_{\theta}(x_t|c)\nabla_{x_t}(f(x_t)\cdot g(c)),
\end{equation}
where $f(x_t)\cdot g(c)$ is the dot product of the CLIP image embeddings and text embeddings. The diffusion process will then perturb the mean of a timestamp with the gradient of the CLIP guided function. 
Although this is useful for enhancing the performance, it increases the training requirement and decreases the image variation. Therefore, classifier-free guidance is also implemented to produce a better performance in inference tasks.

Imagen  \cite{Imagen} is a major work developed on increasingly larger language models, and is regarded as one of the leading works for image fidelity. Their key hypothesis, other than providing improvements to the U-Net architecture, is in regards to the benefits of scaling language models, instead of directly scaling the size of the U-Net. Imagen investigated this through a T5-XXL \cite{T5} encoder which is tasked with mapping the input tokens into a sequence of embeddings. This then gets scaled up using two super resolution layers \cite{Imagen} from a \(64\times 64\) image to a \(1024\times1024\) image. These super resolution layers are modified to be aware of the noise added to the diffusion models through a technique known as noise-level conditioning  \cite{Imagen} to improve the sampling quality. This noise level conditioning is dependent on a cascaded diffusion model structure which is sourced from  \cite{cascaded_diffusion} and works by selecting augmentation values randomly during training and conditioning the diffusion model on this level. Moreover, there are a number of techniques they have to employ from relevant literature in order to tune their diffusion models. They adopt methods from such as classifier free guidance  \cite{Classifier-free_diffusion}, which is a technique that uses gradients from pre-trained models to improve the sample quality of the images. Yet, a diffusion model produced with classifier free guidance  \cite{Classifier-free_diffusion} can suffer from image fidelity issues due to the high guidance weights, and thus they have to make use of thresholding techniques to limit the effect of them.

Their work also introduces many small improvements to the U-Net architecture targeting the inference time, convergence speed and memory efficiency through their efficient U-Net structure. This involves integrating the work from  \cite{Score_based_DDPM, image_refinement} to scale the U-Net skip connections by a constant value to accelerate convergence. In order to reduce the memory cost, Imagen prioritises the lower model parameters with lower resolutions as they contain more channels and thus allow them to increase the capacity of the model. As an attempt to improve inference speed, they reverse the order of down-sampling and up-sampling operations within their blocks such that the down-sampling occurs before the convolutions and the up-sampling occurs after the convolutions. 

Their final text-to-image diffusion model consists of 2B parameters, and \(600\)M/\(400\)M parameters for the two super resolution models, respectively. Using such a large model, they surprisingly did not find any over-fitting issues, and concluded that further training could be beneficial. They also pointed out that the text encoder size, rather than the U-Net's size, is more paramount for image-text alignment and fidelity.
\subsubsection{Prior-decoder TTI}
With the impact of CLIP model that connects text and image, there has been research on a multi-model diffusion process that separates the generation of text embeddings and image embeddings. DALLE-2 \cite{Dalle2} extends the work of GLIDE \cite{Glide} and introduces prior component into their diffusion TTI generation model.

DALLE-2 takes CLIP \cite{Clip} as the text encoder which encodes a text caption into latent space. The prior is an autogressive Transformer or diffusion model that transfers the CLIP text latent space into the CLIP image latent space. When training the autoregressive prior, Principal Component Analysis (PCA) \cite{PCA,SAM} is applied for dimensionality reduction to induce a more efficient training process. When training the diffusion prior, instead of using the noise prediction model \cite{DDPM}, DALLE-2 calculates mean square error (MSE) directly between the noise embedding, $z_i$, and the target embedding, $z_0$. Both of these embeddings end up producing similar results. The decoder is also a diffusion network that converts the CLIP latent space into an image.
DALLE-2 applies a similar strategy as mentioned in denoising diffusion probabilistic models. In the decoder diffusion model, CLIP embeddings and text captions are concatenated into the diffusion model as the guidance. To enable classifier-free training, they randomly drop the text caption roughly half of the time and set the CLIP embeddings to 0-10 percent of the total training time.
In the prior diffusion model, they adopt a method by generating two image embedding samples from  the diffusion model and choose the one that has the highest dot product with the CLIP text embedding. Unlike the traditional diffusion model that predicts the noise distribution, DALLE-2 predicts the un-noised image directly within an arbitrary timestamp where the loss function is based on the mean squared error (MSE) with the true image distribution.

Despite the misclassification that occurs in the CLIP model, such as the notorious examples of classifying an apple to an iPod, this shortcoming never really materialised in DALLE-2.
Although DALLE-2 \cite{Dalle2} requires an extra prior model, it separates the training processes between text conversion and image conversion. In this way, the decoder does not take the text embeddings as a compulsory input to enable text-free training.

Based on the similar structure, Shift Diffusion \cite{Corgi} extends the work of DALLE-2 \cite{Dalle2} and improves upon the sample efficiency of the prior model. Instead of sampling from a randomly normal gaussian space, Corgi \cite{Corgi} explores the possible distribution of the image input. Sampling the noise from a random space in the new distribution shortens the distance between the noise in the start timestep $T$, and the target output. Therefore, there are fewer timestamps that are needed compared to the DALLE-2 prior \cite{Dalle2}. Noise sampling from the new distribution is based on clustering on the training data or simply train a new model using a neural network to approximate the data distribution. Selection of the noise of a image-text pair is based on the following calculation using CLIP:
\begin{equation}\label{eq:corgi-noise-sampling}
    c_y =  {\rm argmax}_{1\leq i\leq k} {\rm Sim}(\mu_i, {\rm CLIP}(text)),
\end{equation}
where $\mu_i$ denotes the mean values of $noise_i$ from the noise collection, $c_y$ is the noise index of selected Gaussian whose $\mu_i$ has the highest cosine similarity with the target image-text pair.
\begin{figure*}[htp] \centering{
\includegraphics[scale=0.55]{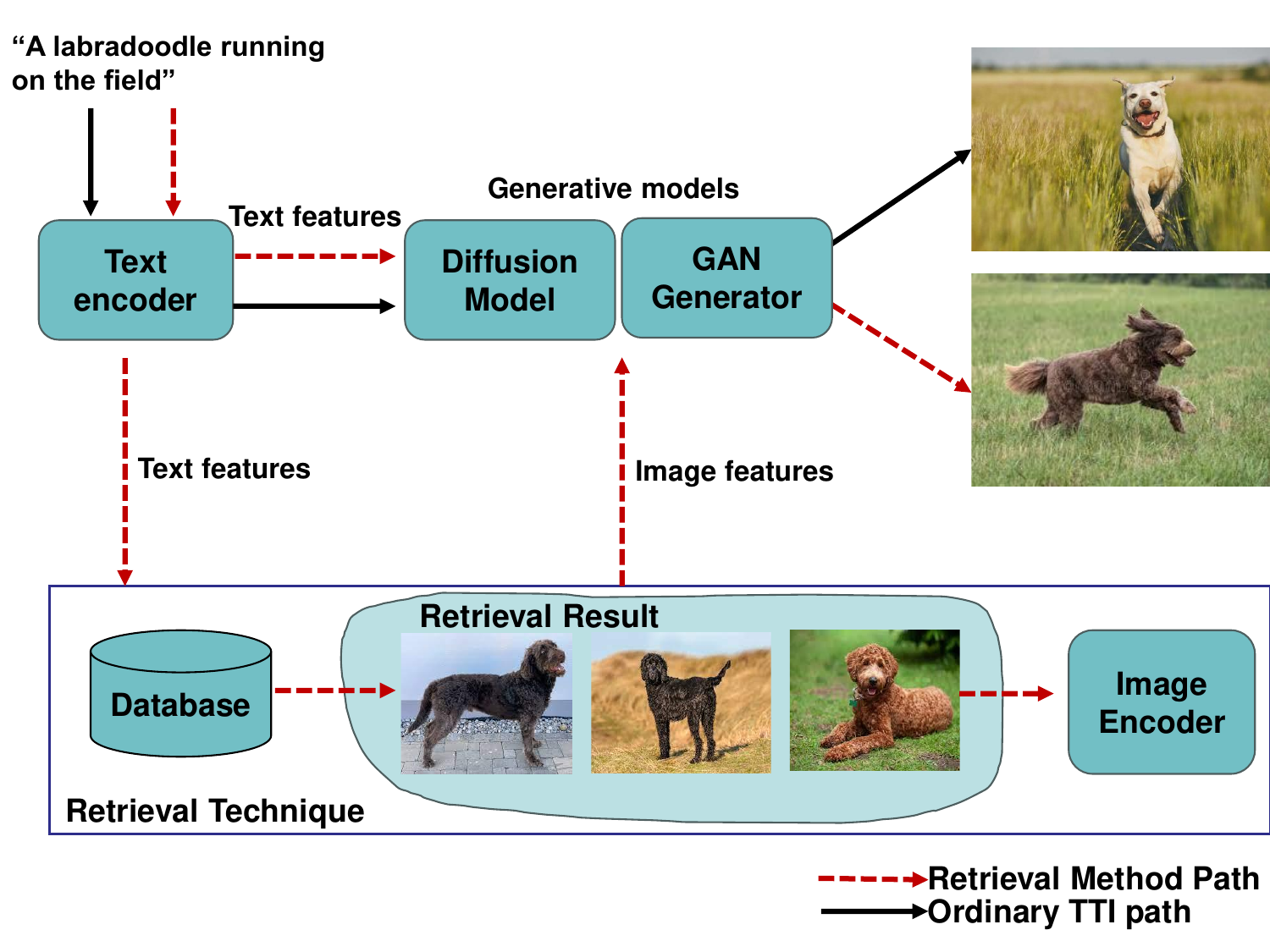}}
\caption{A comparison over TTI model pipeline with and without retrieval technique, where the retrieval technique provides extra visual features for guidance, such techniques can be used for any generative models, examples of GAN and Diffusion method are shown in the figure.}
\label{fig:retrieval}
\end{figure*}
\subsubsection{MOE diffusion}
Refinement of noise prediction within diffusion TTI models stands to gain further refinement through the application of the mixture-of-expert structure \cite{Moe}, thereby offering an avenue for effectively denoising images in different stages of the diffusion process. This proposition aligns with the pioneering work of ERNIE-ViLG 2.0 \cite{Moe_diffusion} and Ediff \cite{ediffi}, both of which underscore the different roles that diffusion model focuses during distinct timesteps.

In the initial stages of the diffusion process, wherein the input resembles random noise to a considerable extent, the diffusion model's objective is to synthesize image intricacies in relation to this noise. This synthesis relies on the interplay of attention mechanisms with text embeddings. Conversely, as the input converges towards the target image, the diffusion model's orientation shifts. It effectively "ignores" the textual guidance and pivots towards the enhancement of image quality itself.

In light of this bifurcation, a novel strategy emerges. The discrete stages of the diffusion process are inherently distinct, prompting the exploration of a multi-network approach. This approach envisions the training of multiple diffusion networks, each with a distinct expertise calibrated to denoise inputs within a particular stage. This differentiation leverages the expertise of each network to optimize denoising effectiveness and bolster the overall performance of the TTI model.

\subsubsection{Retrieval augmented diffusion models}

Diffusion models have seen significant advances through combining their unique U-Net architectures with existing techniques  \cite{Cascaded_u_net,Imagen}. 
One such prominent example is the usage of a retrieval method \cite{Retreival_diffusion,KNN_Diffusion,retrieval-TTI,memorydrivenTTI,ReImagen} that sets up an external explicit memory dedicated for the TTI models. Retrieval models  \cite{Deepmind_retro,KNN_NLP,realm,Token_retreival} have achieved state-of-the-art results in the past by using this external memory to circumvent adding an obnoxious amount of parameters. There are multiple approaches on how to achieve this. Popular ideas include changing the formulation of a standard parametric model into a semi-parametric one conditioned with the external memory or database \cite{Retreival_diffusion}. Other approaches do not use an external retrieval memory but instead use part of the training data to avoid significant memory overhead. However, these latter approaches risk limited generalization capabilities due to their reliance on restricted datasets.

The retrieval technique has been applied to diffusion TTI models through a number of papers  \cite{Retreival_diffusion,retrieval-TTI,ReImagen}, with a common approach demonstrated in Fig~\ref{fig:retrieval}. Retrieval-diffusion \cite{Retreival_diffusion} achieves this through augmenting the model with the latent CLIP \cite{Clip} space of a fixed set of images. The retrieval model is semi-parametric, encoding the non-trainable weights, \(\epsilon_{k}\), as a lookup. The dataset for the retrieval model, defined as \(D\), is where the images are sampled from. The goal of this lookup is to obtain the $k$-nearest neighbour subsets of the dataset based on the input query. This is extremely similar to the work by  \cite{KNN_Diffusion}. The trainable parameters, \(w\), take the results of the query, which are comprised of image patterns, in order to generate the scenes. They could technically provide the results of this query to aid the model by sampling the images directly to update the trainable parameters. However, this would obviously involve significant unknowns and be strained by the dimensionality of the images. Thus, they make use of conditioning in order to reduce the dimensionality by encoding them in a lower dimensional space. This was achieved by using a fixed CLIP image encoder to convert all these examples onto a lower dimensional manifold \(p_{w} (x | \cdot)\). The sampling strategy \cite{Retreival_diffusion} can be expressed as:
\begin{equation}
p_{w, D_{\epsilon_{k}}}(x) = p_{w}(x | \{y | y \in M_{D}^{(k)}\}).
\end{equation}

The database is thus used to approximate a distribution, \(p(x)\), which they will sample queries from to build the \(k\)-nearest neighbours, \(x \sim p(x)\). Since these are encoded in the latent space of CLIP, they use cosine similarity within this image feature space as the metric to search for nearest neighbours, \(\epsilon^{Train}_{k}(x, D)\). When training is completed, the database is replaced with a specific dataset of a certain style. Hence, through the use of retrieval, they are able to apply quite flexible ad-hoc stylisation to their models merely by switching out different stylised datasets. Re-Imagen \cite{ReImagen} is another updated version for Imagen that also takes the strategy of retrieval diffusion and achieves remarkable FID gain in the most used COCO dataset. The same retrieval technique could be also applied to GAN models \cite{TTI-with-Retreival-Gan,RetrieGAN2} and has achieved considerable results with external memory matching semantic relationship between texts and images.
\subsubsection{Latent space diffusion TTI}
Compressing images into low-dimensional latent space representations effectively reduces computational complexity while retaining generation capability in diffusion models. For example, latent diffusion model (LDM) \cite{stable_diffusion} works by mapping the input data to a latent space 4-16 times smaller than the pixel space for the diffusion process. This is performed with the motivation that image details are derived from a very low-level representation of the inputs rather than a richer one informed by the position of a pixel in the context of the image. This method keeps the perceptually relevant details of the image and is able to efficiently sample during the reverse process. Therefore, it could greatly reduce the computation time for both training and inference while ensuring the quality of the sampled image. 

To compress the image into a latent space, LDM adopts an auto-encoder that maps the image data into a lower-dimension representation with equivalent perceptual meaning. Given an image \(x\) with size $H\times W\times3$, the encoder will construct it to the latent variable \(z\) with dimension $h\times w \times c$ by a down-sampling factor \(f\) where $f=\frac{H}{h} = \frac{W}{h}=2^m$. The decoder is trained to reconstruct the latent space variable back to the pixel space, in a form that $\hat{x} = D(E(x))$. Perceptual loss and adversarial loss are introduced during training to enforce local realistic and avoid blurred images.

When fixing the auto-encoder, the diffusion model could be trained on the latent space with the loss function rewritten as follows:
\begin{equation}\label{eq:latent-space-loss-function}
    L = E_{\mathcal{E}(x),\epsilon\sim\mathcal{N}(0,1),t}[\|\epsilon-\epsilon_\theta(z_t,t)\|].
\end{equation}
Besides the above contribution, in order to augment the diffusion process with conditional information, stable-diffusion implements a cross-attention mechanism \cite{Transformer} by projecting all the conditional inputs to a domain-specific latent representation using domain-specific neural models $\tau_\theta(y)$, where $y$ could be any conditional information such as text prompts, semantic map, and images. Attention \cite{Transformer} is then performed in each layer of the U-Net \cite{U-net} backbone, with:
\[Q = W_Q^{i} \cdot \phi_i(z_t),\ K=W_Q^{i} \cdot \tau_\theta(y),\ V=W_V^{i} \cdot \tau_\theta(y),\]
given a U-Net layer \(i\) and intermediate result of noised image \(\phi_i(z_t)\). The conditioning-processing network is trained jointly with the U-Net, using a slightly modified loss objective as follows:
\begin{equation}\label{eq:joint-unet-ldm-loss}
      L_{LDM} = \mathcal{E}_{\epsilon(x),y,\epsilon\sim\mathcal{N}(0,1),t}[||\epsilon-\epsilon_\theta(z_t,t,\tau_\theta(y)||].
\end{equation}

LDM \cite{stable_diffusion} works in a more efficient space with less imperceptible details and more semantics of the data. Besides, it adopts a neural network to process conditional information rather than simply concatenating it into the U-Net backbone. This processing stage makes the conditioning more domain-specific and performs better in guiding the denoising direction for the diffusion model. All these advantages have helped LDM to produce impressive results. LDM is further improved with enlarged training dataset and more varied pre-trained model selections, known as stable-diffusion \cite{stablediffusionweb}, which has become one of the most popular choices for image generation. 

Vector-quantized diffusion (VQ-diffusion) \cite{VQ-diffusion} is another latent space model. Without encoding images with an autoencoder, VQ-diffusion adopts the VQ-VAE \cite{VQ_GAN} structure to reduce the image size into a discrete space to perform the diffusion process. VQ-diffusion introduces two techniques to perturb the latent space data. One is randomly masking the latent space value and the other is replacing one of the value to another randomly discrete value. Besides, VQ-diffusion also implements discrete diffusion algorithm that enables the diffusion model to perform the forward process and discrete process with discrete inputs. With all the contributions, VQ-diffusion is able to generate images efficiently in a much compressed discrete latent space.

In contrast to the majority of recent text-to-image generation methods, there is a significant appeal in creating more models that can be realistically and reliably run on commodity GPU hardware, which is a philosophy that latent models have pushed forth.
\begin{table*}[h]
\begin{center}
\caption{Comparison of different dataset commonly used for computer vision, from small scale image-only dataset to large-scale image-text dataset.}
\begin{tabular}{l|cccc}
\toprule
Public Dataset & Size & Class Label & Captions &Description\\
\midrule
CIFAR10 \cite{Cifar10} & 60K &  10& \ding{55} & Small-scale images with 10 categories.\\
LSUN-Church \cite{LSUN} & 120K & \ding{55}&\ding{55} & Images with church scene.\\
LSUN-Bedroom \cite{LSUN}&3M & \ding{55} & \ding{55} &Images with bedroom scene.\\
ImageNet \cite{Imagenet} & 140M & 1000&\ding{55} & Manual-annotated dataset with 1000 non-overlapped classes.\\
CC3M \cite{CC3M} &3.1M & \ding{55}& 1:1 & Auto-captioned real-world dataset.\\
CC12M \cite{CC12M} & 12M & \ding{55} & 1:1& More diverse image collections based on CC3M.\\
COCOS \cite{COCO} & 120K & 80 & 1:5&Real-world images with 80 object classes.\\
LAION400M \cite{Laion-400M} & 400M  & \ding{55} & 1:1 & Large scale dataset with web-crawled images.\\
LAION5B \cite{LAION-5B} & 5B  & \ding{55} & 1:1& Enlarged Laion dataset with multiple languages.\\
\bottomrule
\end{tabular}
\label{dataset-comparison-table}
\end{center}
\end{table*}
\begin{table*}[h]
\begin{center}
\caption{List of evaluation metrics and benchmarks for text-to-image models, where FID is the most commonly used metrics that covers image fidelity and diversity. While benchmarks covers thorough comparisons, they necessitate human evaluations, leading to increased time and effort investments.}
\begin{tabular}{lcccccc}
\toprule
Metric & Fidelity & Diversity & Text alignment & Human Evaluation& Language\\
\midrule
IS\cite{IScore} &  \ding{51}& \ding{51} & \ding{55} &\ding{55}& -\\
FID\cite{FID-Score} & \ding{51} &\ding{51}& \ding{55} &\ding{55}& -\\
R-precision\cite{AttentionGAN} &  \ding{55}&\ding{55} & \ding{51} &\ding{55}& -\\
CLIP-Score\cite{clipscore} & \ding{55} &\ding{55} & \ding{51} & \ding{55} & English\\
UniBench \cite{UniBench}& \ding{51} & \ding{55} & \ding{51} &\ding{51}& Chinese, English\\
PaintSKills \cite{paintskills} & \ding{51} & \ding{55} & \ding{51} &\ding{51}& English\\
DrawBench \cite{Imagen} & \ding{51} &\ding{55} &\ding{51} &\ding{51}& English\\
PartiPrompts \cite{Parti} &\ding{51} & \ding{55} &\ding{51} & \ding{51} & English\\
\bottomrule
\label{Tab:Tcr}
\end{tabular}
\end{center}
\vspace{-1mm}
\end{table*}
\subsubsection{Sketch guided control}
Another remarkable technique to generate images that meet users' expectation is the introduction of additional pre-defined input parameters, which enable users to steer the diffusion process with greater precision toward their envisioned image outcomes. By providing expected image specifications such as segmentation maps, object locations, and edge maps, users can now exert precise control over the generated images \cite{sketch_based_diffusion,ControlDiffusion,spatext}. Spatext \cite{spatext} extracts image segments during training process through a pre-trained segmentation model,  and their corresponding CLIP image embeddings are computed. These image embeddings are then strategically stacked to mirror the shape and location of their respective segments within the original image, thus forming a comprehensive spatio-textual representation. This representation is then introduced as conditional input to the U-Net. This innovative training methodology empowers the diffusion model to generate images that closely adhere to the characteristics of the spatio map. During the inference process, users provide image sketches accompanied by local prompts that describe segmented areas. These prompts are processed through a prior model \cite{Dalle2} to convert their text embeddings into image embeddings. Subsequently, these embeddings are stacked and mapped onto the spatial space where they align with the sketch's shapes and locations. This approach enables Spatext \cite{spatext} to generate high-fidelity images, rich in object details that precisely match the user-provided shape, location, and color specifications.

Controlnet \cite{ControlDiffusion}, meanwhile, extends the current stable-diffusion structure with replicate U-Net layers sharing the same weights with the original U-Net. Controlnet \cite{ControlDiffusion} receives extra pre-defined image information such as background colors and edge maps. During training, these additional image information are channeled into the replicate U-Net. Given that there is a convolution layer with zero weights attached to the end of each new U-Net layer, the initial training phase does not compromise the performance of the pre-trained stable diffusion model. However, as iterations progress, the new U-Net learns to leverage the extra image information to enhance predictions from the stable diffusion model. After short iterations of finetuning, the new U-Net is able to help guide the diffusion process towards the sketch and domain pre-defined by users.

\section{Comparison of approaches}\label{comparisons}
\begin{table*}[h]
\begin{center}
\caption{A comparison of the text to image methods discussed highlighting their date published, model configuration and evaluation results. For the model type, green dot refers to the GAN model TTI, blue dot refers to the autoregressive TTI and orange dot indicates the Diffusion TTI. For evaluation metrics, IS and FID score are provided under the evaluation of MSCOCO dataset in a zero-shot fashion. The last column provides the specific model size in scale of Million(M) or Billion(B); $\star:$ no zero-shot results found, use standard results instead.}
\begin{tabular}{lcccccccc}
\toprule
Method & Date & Model Type  & Data Size & Open Source & IS evaluation & FID evaluation & Model size\\
\midrule
AttnGAN \cite{AttentionGAN} & 11/2017 & \tikz\draw[green,fill=green] (0,0) circle (.5ex);& 120K&  \ding{55} & 20.80 & 35.49 $\star$ & 13M \\ 
StyleGAN \cite{StyleGan} & 11/2017 & \tikz\draw[green,fill=green] (0,0) circle (.5ex);& 120K&  \ding{55} & 20.80 & 35.49 $\star$ & - \\ 
Obj-GAN \cite{object_gan} & 09/2019 &  \tikz\draw[green,fill=green] (0,0) circle (.5ex); & 120K&\ding{51} & 24.09 & 36.52 $\star$& 34M\\
Control-GAN \cite{ControlGan} & 09/2019 &  \tikz\draw[green,fill=green] (0,0) circle (.5ex); & 120K&\ding{51} & 23.61 & 33.10 $\star$& -\\
DM-GAN \cite{DMGAN}& 04/2019 &\tikz\draw[green,fill=green] (0,0) circle (.5ex); & 120K & \ding{51} & 32.32 & 27.34 $\star$&21M\\
XMC-GAN \cite{XMC-GAN} & 01/2021 & \tikz\draw[green,fill=green] (0,0) circle (.5ex); & 120K & \ding{55} & 30.45 & 9.33 $\star$& 90M\\
LAFITE \cite{lafite} & 11/2021 & \tikz\draw[green,fill=green] (0,0) circle (.5ex);&-&\ding{51} & 26.02 &26.94 & 150M \\
Retreival-GAN \cite{TTI-with-Retreival-Gan} & 08/2022 &  \tikz\draw[green,fill=green] (0,0) circle (.5ex); & ~120K &\ding{55} & 29.33 & 9.13 $\star$& 25M \\
GigaGAN \cite{GigaGan}& 01/2023&\tikz\draw[green,fill=green] (0,0) circle (.5ex); &-&\ding{55}& -&10.24 & 650M \\
GALIP \cite{galip} & 03/2023 & \tikz\draw[green,fill=green] (0,0) circle (.5ex); &3M-12M& \ding{51} & - & 12.54 & 240M\\
DALLE \cite{DALLE} & 02/2021 & \tikz\draw[blue,fill=blue] (0,0) circle (.5ex);& 250M& \ding{55} & - & 27.5 & 12B\\
Cogview \cite{Cogview}& 06/2021 & \tikz\draw[blue,fill=blue] (0,0) circle (.5ex);&300M&\ding{51} & - &27.1 & 4B\\
Make-A-Scene & 03/2022 & \tikz\draw[blue,fill=blue] (0,0) circle (.5ex);&35M&\ding{55} & - & 11.84 & 4B\\
Cogview2 \cite{cogview2} & 05/2022 & \tikz\draw[blue,fill=blue] (0,0) circle (.5ex);&300M&\ding{51} & - &24.0 & 6B\\
PARTI-350M \cite{Parti} & 06/2022 & \tikz\draw[blue,fill=blue] (0,0) circle (.5ex); &  $\sim$1000M&\ding{55}& -&14.10 & 350M \\
PARTI-20B \cite{Parti} & 06/2022 & \tikz\draw[blue,fill=blue] (0,0) circle (.5ex); &  $\sim$1000M&\ding{55}& -&7.23 & 20B \\
DALLE-mini \cite{dayma_2021_dalle_mini} & 07/2021 & \tikz\draw[blue,fill=blue] (0,0) circle (.5ex); & 250M & \ding{55}&-& - & $\sim$500M \\
MUSE-3B \cite{muse} & 03/2023 & \tikz\draw[blue,fill=blue] (0,0) circle (.5ex); &  $\sim$1000M&\ding{55}& -&7.88 & 7.6B \\
GLIDE \cite{Glide} & 12/2021 & \tikz\draw[orange,fill=orange] (0,0) circle (.5ex);  & 250M & \ding{51} & -&12.24 & 5B\\
VQ-diffusion-F \cite{VQ-diffusion}& 11/2021& \tikz\draw[orange,fill=orange] (0,0) circle (.5ex);& $>$7M &\ding{51}& - &13.86 $\star$ & 370M\\
DALLE-2 \cite{Dalle2} & 04/2022 & \tikz\draw[orange,fill=orange] (0,0) circle (.5ex); & 250M& \ding{55}&-& 10.39 & 5.2B\\
Imagen \cite{Imagen} & 05/2022 & \tikz\draw[orange,fill=orange] (0,0) circle (.5ex);  & $\sim$860M& \ding{55}&-& 7.27  & 7.6B\\
LDM \cite{stable_diffusion} & 08/2022 & \tikz\draw[orange,fill=orange] (0,0) circle (.5ex); & 400M &\ding{51} &30.29& 12.63 & 1.45B\\
eDiff-I \cite{ediffi}&11/2022 &\tikz\draw[orange,fill=orange] (0,0) circle (.5ex); & 1000M &\ding{55}&-&6.95 & 9B\\
Shift Diffusion\cite{Corgi} & 08/2022 & \tikz\draw[orange,fill=orange] (0,0) circle (.5ex); & 900M &\ding{51}&-&10.88&- \\ 
Re-Imagen\cite{ReImagen} & 09/2022 & \tikz\draw[orange,fill=orange] (0,0) circle (.5ex); & 50M&\ding{55}&-& 6.88 & $\sim$8B \\
ControlNet \cite{ControlDiffusion}& 03/2023 & \tikz\draw[orange,fill=orange] (0,0) circle (.5ex);& - & \ding{51} &-&-&$\sim$2.2B\\
\bottomrule
\end{tabular}
\label{method-comparison-table}
\end{center}
\end{table*}
\subsection{Experimental Setups and Selection of Dataset}

The comparative options for training generative models are presented in Table \ref{dataset-comparison-table}. CIFAR10 \cite{Cifar10}, ImageNet \cite{Imagenet}, and LSUN \cite{LSUN} are labelled datasets that categorize images into various scenes and objects. Widely embraced, they form the bedrock for training fundamental generative models in scenarios of unconditional or class-conditional sampling\cite{DDPM,DDIM,DDPM_Beats_Gan}. Additionally, these datasets serve as valuable resources for training image encoders and super-resolution models\cite{stable_diffusion}, integral components of the TTI model framework. However, their applicability is generally not extended to training text-conditioned image generation.

MSCOCO \cite{COCO} contains approximately \(330\)k captioned images and over \(200\)k have class labels. There are in total \(80\) object categories and \(91\) stuff categories and on average 3.5 categories per image. Reflecting diverse real-world scenarios, MSCOCO has evolved into a widely recognized evaluation dataset, assessing model performance across image segmentation \cite{DeepLabv3,PolarMask}, classification, and object detection. Besides, MSCOCO \cite{COCO} offers utility in training small-scale GAN TTI models in preliminary stages, given its compact size and computationally less demanding nature, while still yielding satisfactory results. Although no longer a primary dataset for training large TTI models, MSCOCO's high-quality image-text pairs continue to hold value as a crucial evaluation benchmark. In particular, they serve to rigorously test TTI model performance under zero-shot conditions, offering essential insights into model distinctions and capabilities.

LAION \cite{Laion-400M} emerges as a large dataset encompassing approximately 400 million image-text pairs in its initial iteration, a number that has since escalated to 5 billion \cite{LAION-5B} due to a continuous influx of new images. The image-text pairs are retrieved from random web pages with Common Crawl \cite{commoncrawl} and are filtered by CLIP score to ensure image-text alignment. Beyond the textual descriptions, LAION400M also equips users with CLIP embeddings and KNN indices associated with the image-text pairs. These ancillary resources serve as valuable assets for diverse analytical tasks involving clustering, K-nearest methods, and data subset filtration. This augmented resource environment significantly bolsters the training and inference process of retrieval-augmented TTI models, without the need of excessive time to collect extra data and build external database. 

The Conceptual Captions Datasets \cite{CC3M,CC12M} also constitute substantial resources containing image-text pairs. In pursuit of a harmonious blend of qualities, informativeness, fluency, and learnability in the ensuing captions, a systematic automatic pipeline is employed. This pipeline orchestrates the extraction, filtration, and transformation of potential image/caption pairs, thereby achieving a fine balance in the dataset's attributes.

For the training of large TTI models, an enlarged dataset spanning diverse domains is imperative. This broadens the models' generalizability, enabling them to yield real-world images and effectively operating with zero-shot captions. Notably, the images curated from these large image-text datasets, LAION and Conceptual Captions, stand as premier choices for training large TTI models, encapsulating the requisite diversity to ensure robust performance and broad domain coverage.

\begin{table*}[ht]
\centering
\renewcommand{\arraystretch}{1.2}
\caption{Comparison of the visual result from different TTI models, where we select 4 models and generate 5 images respectively for each model.Text prompts are randomly selected.}
\begin{tabular}{|l|*{5}{>{\centering\arraybackslash}m{3cm}|}}
\hline
\textbf{Model} & \textbf{A peaceful countryside scene with a charming cottage and blooming flowers.} & \textbf{An adorable kitten playing with a colorful ball of yarn} & \textbf{A surreal dreamlike landscape with floating islands and rainbow-colored waterfalls} & \textbf{An underwater world teeming with colorful coral reefs and exotic fish.} & \textbf{A profile photo for a smart, engaging digital assistant.} \\
\hline
GLIDE & \includegraphics[width=3cm,height=3cm]{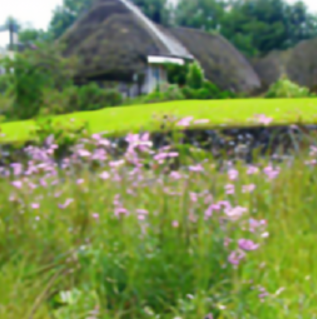} & \includegraphics[width=3cm,height=3cm]{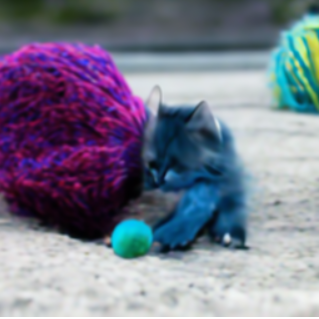} & \includegraphics[width=3cm,height=3cm]{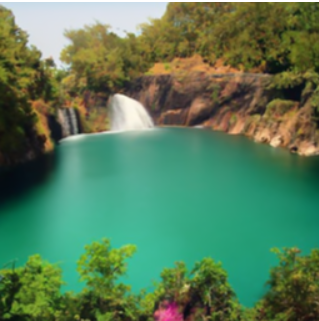} & \includegraphics[width=3cm,height=3cm]{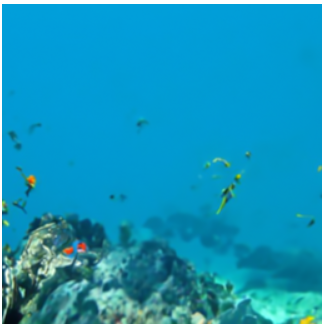} & \includegraphics[width=3cm,height=3cm]{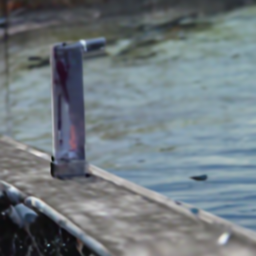} \\
\hline
DALLE-2 & \includegraphics[width=3cm,height=3cm]{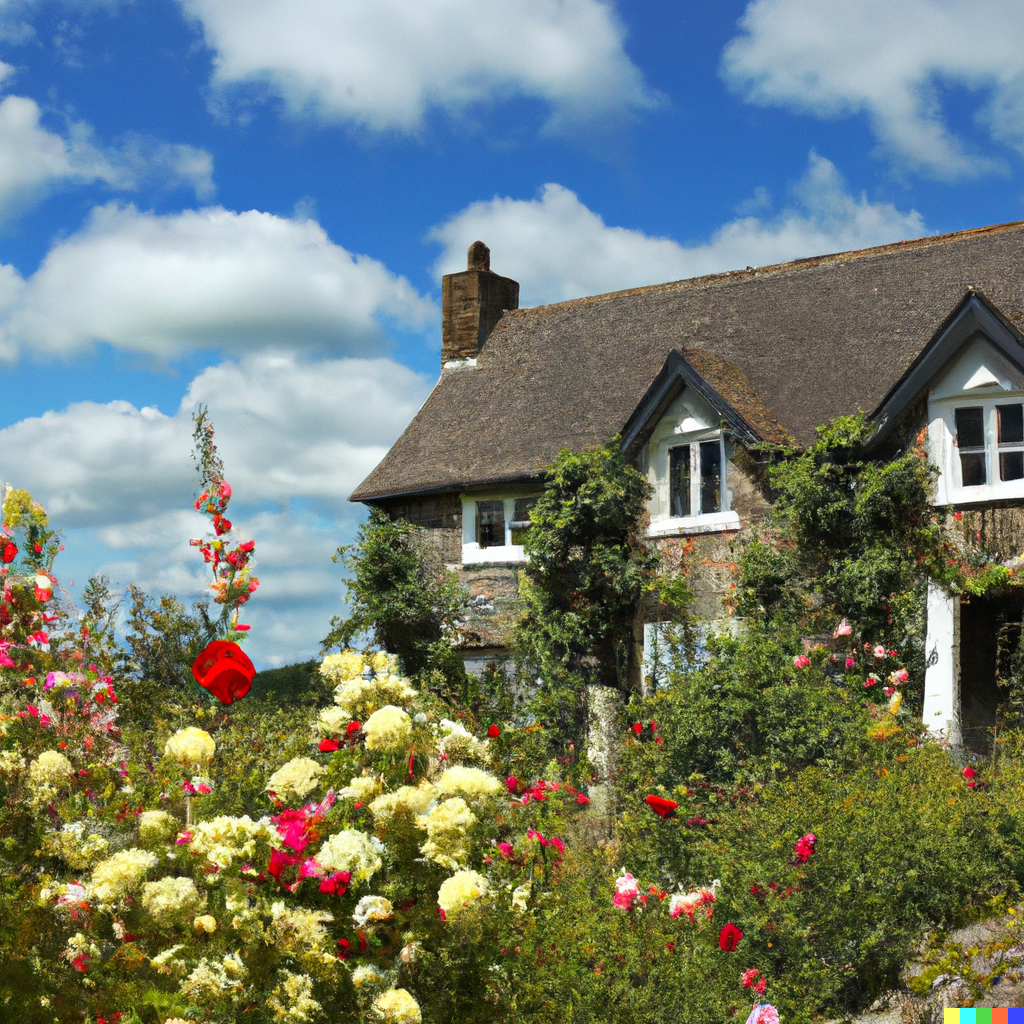} & \includegraphics[width=3cm,height=3cm]{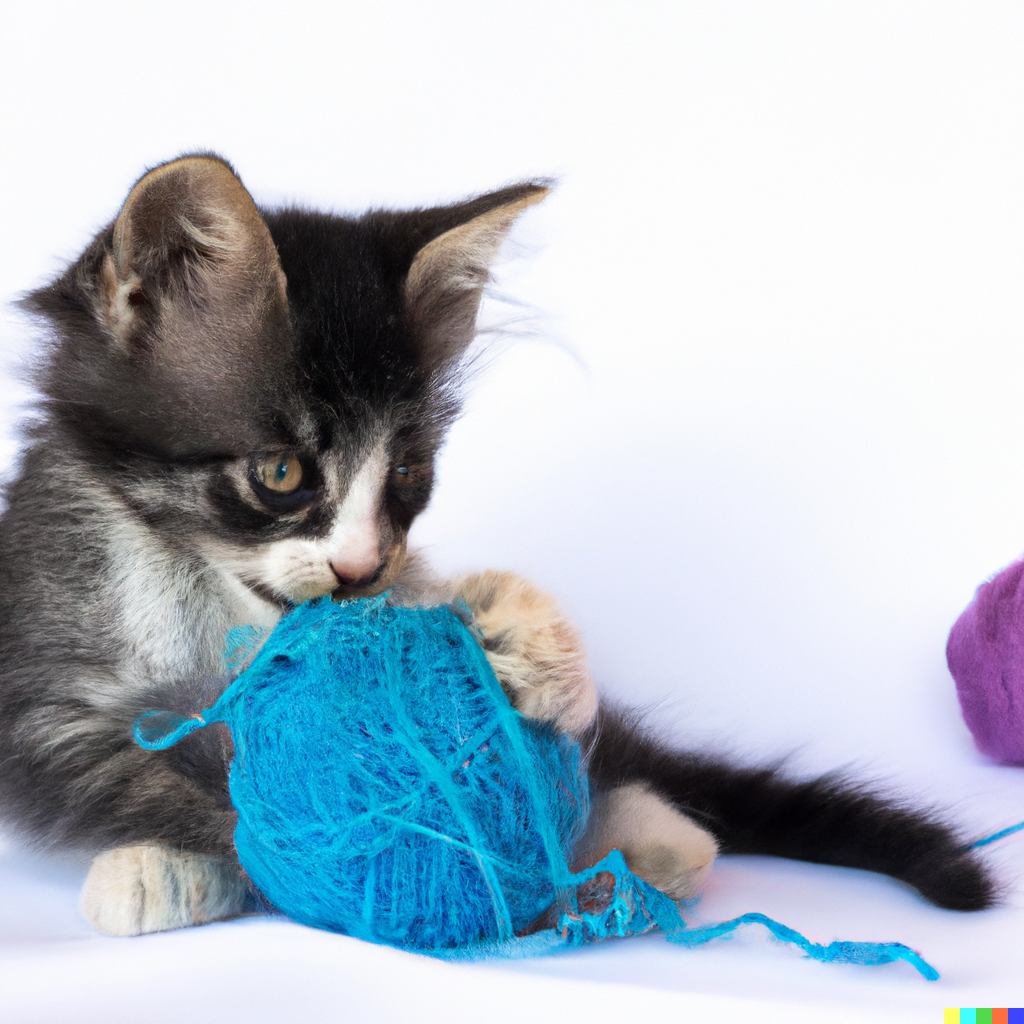} & \includegraphics[width=3cm,height=3cm]{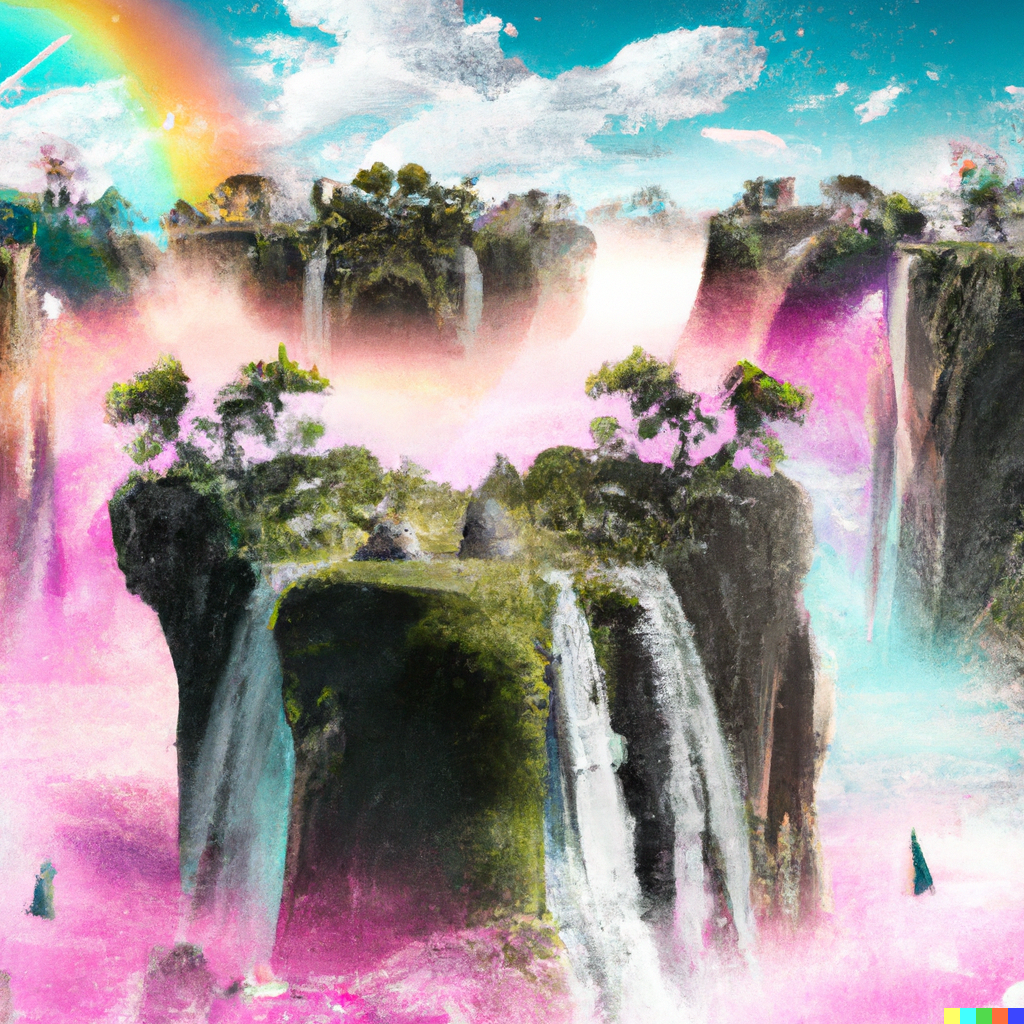} & \includegraphics[width=3cm,height=3cm]{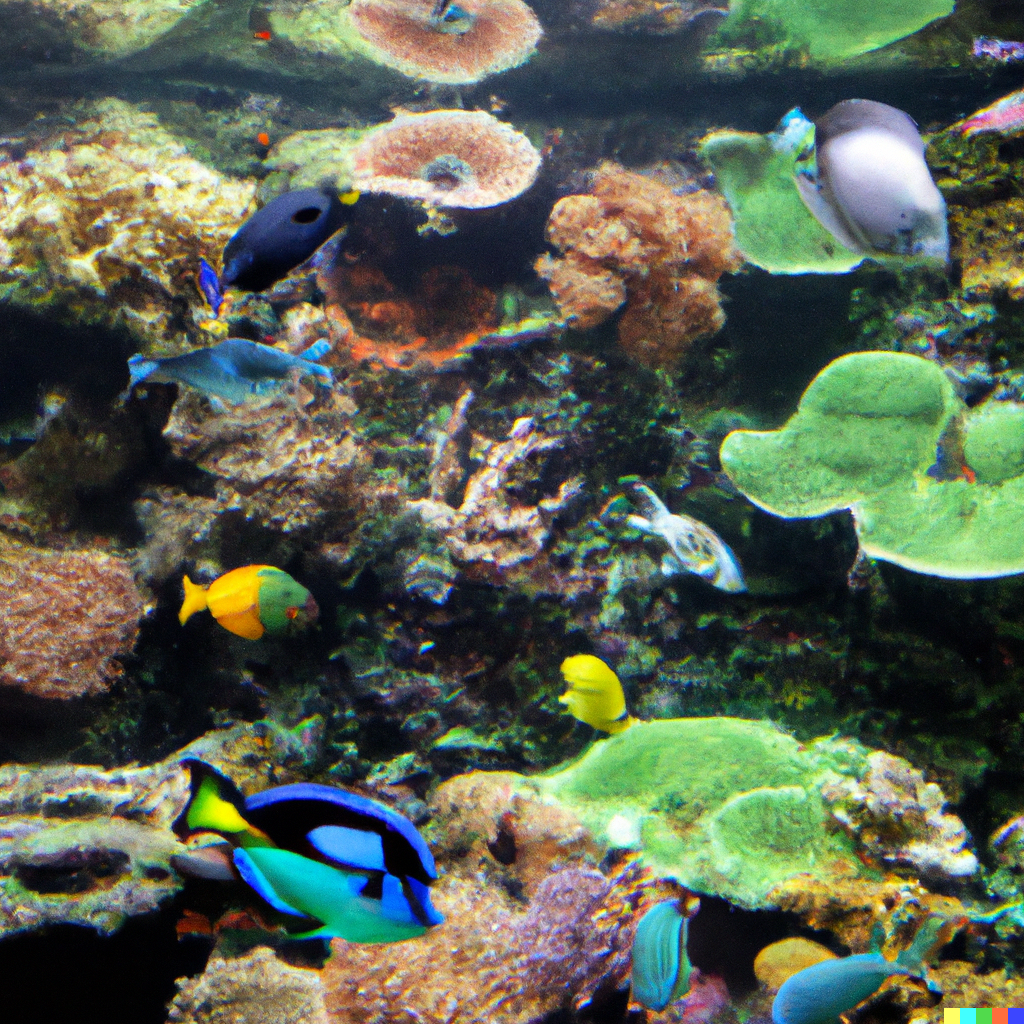} & \includegraphics[width=3cm,height=3cm]{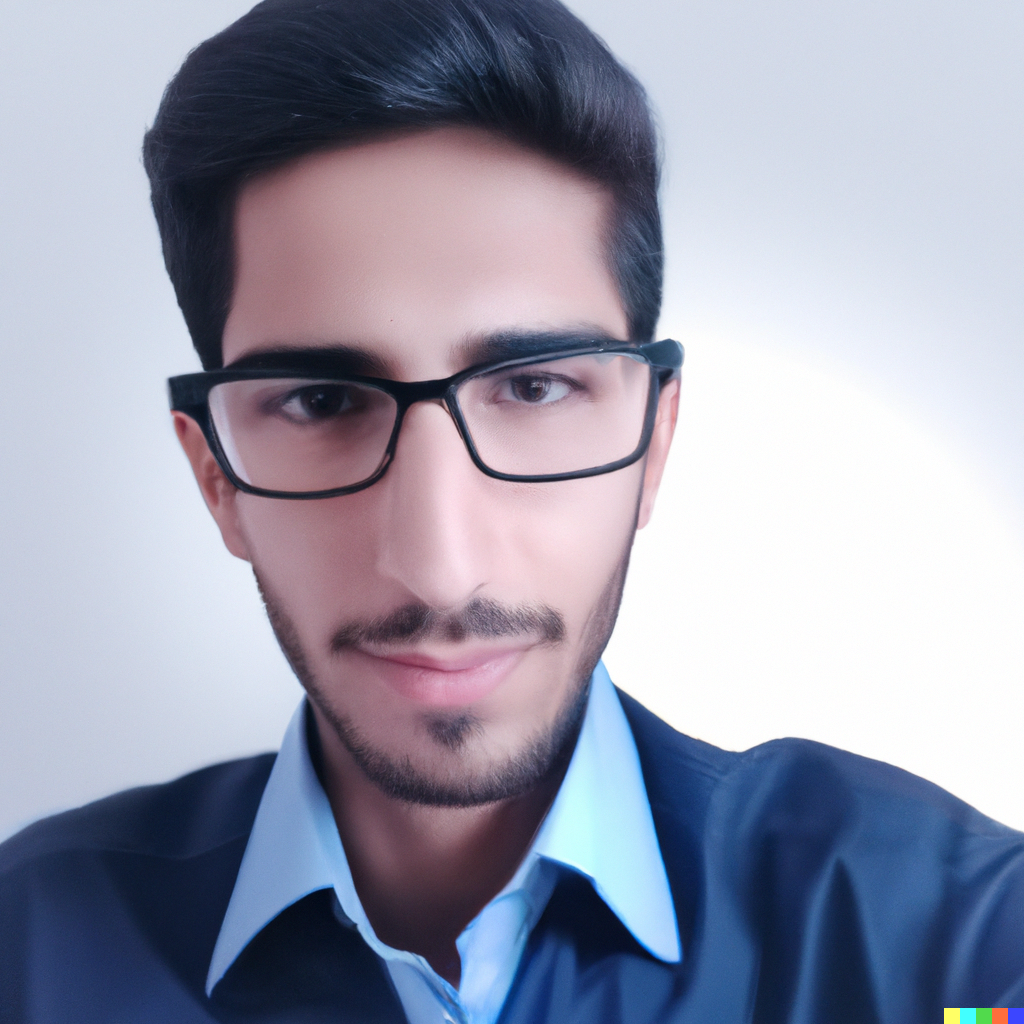} \\
\hline
SD & \includegraphics[width=3cm,height=3cm]{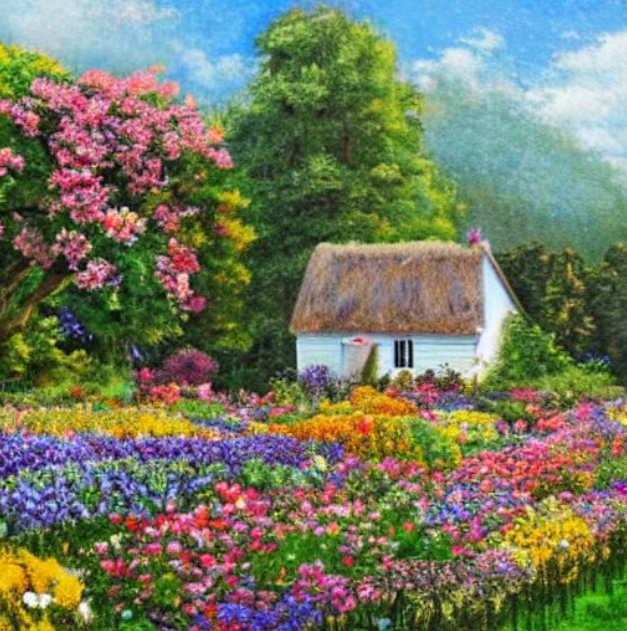} & \includegraphics[width=3cm,height=3cm]{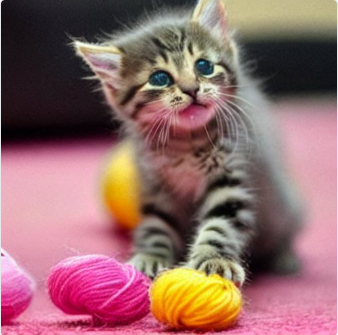} & \includegraphics[width=3cm,height=3cm]{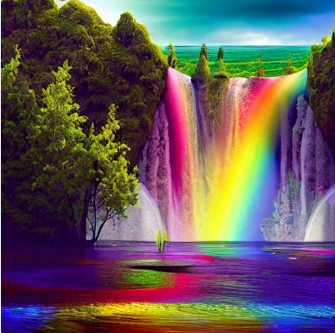} & \includegraphics[width=3cm,height=3cm]{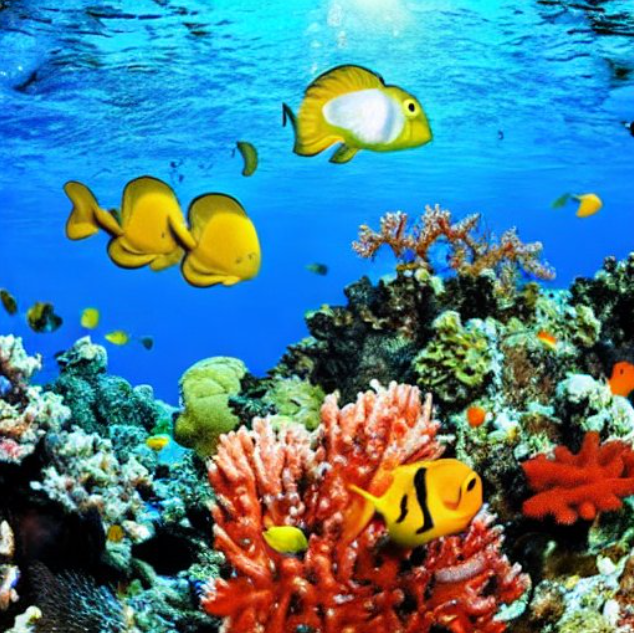} & \includegraphics[width=3cm,height=3cm]{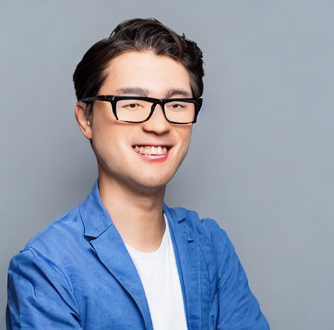} \\
\hline
GALIP & \includegraphics[width=3cm,height=3cm]{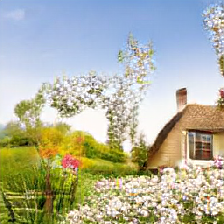} & \includegraphics[width=3cm,height=3cm]{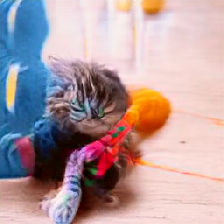} & \includegraphics[width=3cm,height=3cm]{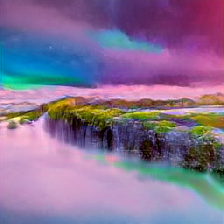} & \includegraphics[width=3cm,height=3cm]{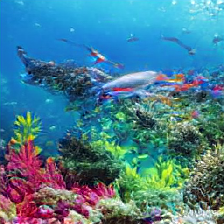} & \includegraphics[width=3cm,height=3cm]{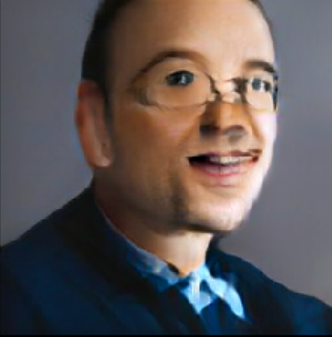} \\
\hline
\end{tabular}
\label{tab:comparison}
\end{table*}
\subsection{Evaluation Metircs and Benchmarks}
Evaluating the performance of TTI models is a critical part of their training process. Researchers have adopted a number of evaluation metrics and benchmarks as a scale of TTI model performance, as shown in Table \ref{Tab:Tcr}. Inception Score (IS) \cite{IScore} is an algorithm that measures the image quality for generative models. It is applied as an evaluation metric for most of the GAN TTI models. Inception score \cite{IScore} uses a pre-trained Inception model \cite{InceptionModel} to predict the conditional label or text distribution \(P(y|x)\). The inception score, as detailed in Eq. (\ref{eq:inception-score}), is calculated through the KL divergence of the conditional distribution and the real distribution:
\begin{equation}\label{eq:inception-score}
    {\rm IS} = \exp(\mathcal{E}_x(KL(p(y|x)||p(y)))).
\end{equation}

The IS reflects the quality and diversity of the synthesised images. It could be maximized when the following two conditions are satisfied:
\begin{enumerate}
    \item The conditional distribution $P(y|x)$ has low entropy for each generated image, implying that the generated image could be clearly identified to a class label by the pretrained model. 
    \item The marginal distribution of generated samples $\int p(y|x=G(z))$ has a high entropy, so the model is able to generate images with diversity.
\end{enumerate}

The drawback of the IS is that it lacks the comparison between the generated images and real-world images, as it is calculated within the pretrained model \cite{InceptionModel}. Therefore, IS is only used in the early GAN TTI models, and a few of the updated Diffusion TTI models. The drawback of Inception Score is resolved by the Féchet Inception Distance (FID) \cite{IS_distance} with more consistency in comparison to real-world samples. The FID score calculation is detailed in Eq. (\ref{eq:fid-score}) \cite{FID-Score}:
\begin{equation}\label{eq:fid-score}
\begin{split}
        d^2((\mathbf{m},\mathbf{C}),(\mathbf{m_W},\mathbf{C_w})) = 
        ||\mathbf{m}-\mathbf{m_w}||^2 \\ + \mathbf{Tr}(\mathbf{C}+\mathbf{C_w}-2(\mathbf{CC_w})^\frac{1}{2}),
\end{split}
\end{equation}
where \(d\) refers to the Féchet Distance \cite{Multivariate_FID}. The values of \(\mathbf{m}, \mathbf{C}, \mathbf{m_w}, \mathbf{C_w}\) are the mean and covariance matrices of the evaluation dataset and generated samples in the last pooling layer of the pre-trained Inception model \cite{InceptionModel}, respectively~ \cite{IScore}.  Besides, FID score is calculated in every mini-batch \cite{FID-Score} during the training process. FID score \cite{FID-Score} absorbs the similarity between synthetic images and real-world images by comparing the image distribution from both output samples and evaluation samples, this ensures the generated images to have high qualities and diversities as the evaluation images. It has become the standard metric to assess model performance for a wide range of GAN, auto-regressive Transformer and Diffusion-based TTI models under MSCOCO dataset \cite{COCO}, as an alternative to the IS.

Despite the ability to involve the real world samples, FID is biased from the generator given the finite number of training samples \cite{FID_InFinity}. The same reason also causes the IS score to be log negatively biased \cite{FID_InFinity}. This problem could be alleviated using FID\(_\infty\) and IS\(_\infty\) \cite{FID_InFinity}. Besides, FID could produce inconsistent results from human judgements \cite{Imporved_FID} with unmatched data-sets between the Inception model and training model.

Beside the well-established IS and FID metrics, there exists a spectrum of other evaluation measures tailored to specific aspects of TTI model performance. The CLIP score, for instance, hones in on image-text alignment, leveraging a pretrained CLIP model for computation. An alternative approach, R-precision \cite{AttentionGAN}, quantifies visual-semantic affinity by ranking retrieval outcomes between extracted image and text features. 

Furthermore, a comprehensive suite of human-evaluation benchmarks aids in TTI model comparison, focusing on aspects like image fidelity and text-image alignment. For example, Unibench \cite{UniBench} selects 200 text prompts in Chinese and English, with 100 simple-scene prompts and 100 complex-scene prompts, to collect raters' preference by the comparison of two model outputs. Imagen \cite{Imagen} introduces DrawBench, featuring 200 prompts across 11 categories, serving as a testing ground for diverse capabilities such as color rendering and object recognition.  PartiPrompts \cite{Parti} elevates the prompt count to 1600, encompassing distinct categories and facets that diverge from DrawBench's scope.

\subsection{Comparison of TTI models}
Table \ref{method-comparison-table} details the model sizes and best evaluation results for a plethora of TTI models. For quantitative analyses, we provide IS and FID results tested on MSCOCO dataset in a zero-shot fashion. While these benchmarks cannot definitively adjudicate the performance of models in the absence of additional human-evaluation benchmarks, they do offer valuable insights into the performance of these models across a broad range of real-world scenarios.

Earlier GAN TTI models exhibit small model parameter counts less than $100M$ threshold. This limited model scale impedes the model performance and its applicability in large dataset training for zero-shot TTI generation \cite{GigaGan}. These limitations have caused poor evaluation results with high FID and low IS even not in a zero-shot environment. Despite their substantial disparity with contemporary TTI models, recent work applying novel model architectures and scaling up model sizes have engendered a staggering boost in GAN TTI performance. These advancements have propelled GAN TTI models to rival with some of the latest Diffusion and Autoregressive TTI models. Notable examples are GALIP \cite{galip} and GigaGAN \cite{GigaGan}, which exhibit impressive FID scores of $12.54$ and $10.24$, respectively, while maintaining a compact model size. Moreover, GAN TTI models have a natural computational advantage, generating images in a singular step, whereas both diffusion and autoregressive Transformers require iterative generation process. This combination of model size and computational efficiency position GAN TTI models as promising contenders for future TTI applications, underscoring their untapped potential.

Autoregressive Transformers typically encompass a higher number of model parameters, driven by the necessity for large language models and multiple attention layers. Recent progressions in architectural designs and the scale up of model sizes have markedly enhanced their performance, culminating in some of the most exceptional results within the domain of TTI generation. Their large training datasets enable further bolsters to the zero-shot capabilities. Notably examples are PARTI-20B \cite{Parti} and MUSE \cite{muse}, whose FID scores are only marginally higher than state-of-the-art Diffusion models.

Diffusion model, on the other hand, have demonstrated their capacity to achieve commendable synthetic outcomes even with moderate model parameters, as evident in the case of LDM \cite{stable_diffusion}. Diffusion models have undergone substantial advancements in recent years, with significant techniques introduced to enhance their performance. Two noteworthy examples are eDiff-I \cite{ediffi} and Re-Imagen \cite{ReImagen}, which yield FID scores of $6.95$ and $6.88$, respectively, leveraging the advantages of MOE and retrieval techniques. These outcomes surpass the performance of all GAN-based TTI models and slightly better than state-of-the-art autoregressive Transformers. However, introduction of these techniques has inevitably increased the memory requirement and computation time,  consequently leading to larger model sizes.

In addition to this, Table \ref{tab:comparison} gives a succinct visual comparison among distinct TTI models. Their effectiveness is assessed through the analysis of five randomly generated text prompts, drawn from ChatGPT-3 \cite{chatgpt-3}. Each prompt encompasses a distinct zero-shot scenario. All models generate outputs that achieve an overall satisfactory quality. Nonetheless, upon closer examination, specific nuances in their performance become evident. For instance, GLIDE \cite{Glide}, representing an early diffusion model, exhibits a minor degree of information loss that occasionally results in misalignment with the provided text prompts. This discrepancy is especially obvious in its last column,  where the text interpretation is severely misaligned. On the other hand, DALLE-2 \cite{Dalle2}, Stable-Diffusion \cite{stablediffusionweb}, and GALIP \cite{galip} showcase exceptional image detailing and vivid color rendition. Despite these achievements, some images within their outputs display blurred objects or unrealistic scenes. It is noteworthy that stable-diffusion \cite{stable_diffusion} and GALIP \cite{galip} encounter  challenges when synthesizing images featuring faces, as exemplified by the text prompt "A peaceful countryside scene with a charming cottage and blooming flowers."
\section{Future outlook} 
\subsection{Limitations and Potential Improvement of Current TTI Models}
While various types of TTI models have showcased their respective strengths and limitations, there remains significant potential for the development of models that mitigate these shortcomings. The goal is to create resource-efficient and computationally expedient models capable of generating high-fidelity images. Such advancements would extend the applicability of these models to a broader range of local computing hardware and even mobile devices. 

Currently, a trade-off between sampling efficiency and image quality persists across TTI models. State-of-the-art TTI generation models typically require around 50-200 iterative steps to synthesize high-quality images. Autoregressive Transformers, while achieving impressive results, demand substantial memory resources and employ an iterative generation process to predict image tokens. GAN-based TTI models, although promising, still exhibit a minor performance gap when compared to the leading autoregressive Transformers and Diffusion models.

Given these ongoing limitations, there remains a need for continued research efforts aimed at addressing the shortcomings of all types of TTI models. One avenue worth exploring is the investigation of model architectures and algorithms that facilitate TTI generation with minimal steps, such as GAN models \cite{galip,GigaGan}, non-autoregressive Transformers \cite{muse}, and Diffusion models enhanced by recent distillation techniques \cite{progressive,linearflow} and advanced ODE solvers \cite{dpmsolver}. By advancing these models to overcome their existing constraints, researchers can contribute to the realization of more versatile and efficient text-to-image generation techniques.
\subsection{Beyond TTI: Embrace the Era of AIGC}
\subsubsection{TTI applications under the influence of AIGC}
The development of AI-generated content (AIGC) has brought significant influence in many social and research aspects. New AIGC models have made it possible for people to access a vast amount of information on various topics, which will be beneficial to a wide range group of people and occupations. With the help of ChatGPT \cite{chatgpt4} generated content, people can quickly retrieve information on a wide range of subjects without significant expertise or having to spend time researching and reading through a vast amount of material. For writers, ChatGPT can also help to generate new ideas for stories, characters, and plotlines. Using these generative models, it is even possible to suggest titles, opening sentences, and themes for writers to explore with. Provided that more integration of AIGC models to the industry fields is happening, AIGC can be projected to see endless potential to boost the efficiency and productivity for people with their works. 

Following the trend of AIGC, text-to-image models also expect the possibility to support varied generation tasks to apply in a wide range of general or industrial aspects. Some TTI models have enabled multiple generative tasks other than TTI generation. For example, stable-diffusion \cite{stable_diffusion} not only supports TTI generation, but also enables image in-painting and super-resolution. Imagen \cite{Imagen} and GigaGAN \cite{GigaGan} both enable style transfer given image inputs. Such features potentially have a significant impact on both creativity and productivity on the fields of art creation and fashion design, with artists and designers seeking digital art and illustrations more easily and quickly than ever before. It posits the potential of providing these creators with the ability of creating art in different styles, including realistic, abstract, and surrealistic. This will also allows them to explore new possibilities and push the boundaries of what is possible. With further development of TTI models, we can also expect more plugin and software support such as Midjourney as well as integration to a broad range of tasks that suffice human needs, leading to a refined ecosystem with close connections to our daily life.

\subsubsection{Extend the task of TTI}
TTI techniques, with their generative models capable of generating new contents based on user inputs, are closely associated with other generative tasks that employ similar model architectures and algorithms. There has been development for video generation \cite{video_diffusion,videogpt,Improved_diffusion_Long_video,Flexi_diffusion_video_gen,make-a-video,videoimagen,videocomposer,probadaptation,makeyourvideo,controlavideo,genlvideo,mmdiffusion} with the involvement of diffusion techniques, which produces a sequence of image frames. Instead of using the original U-Net structure, video-diffusion model \cite{video_diffusion} implements 3D-UNet structure with each block representing 4D tensor to process fixed-length time-framed images. Current TTI techniques have also been successfully applied in diffusion video generation, such as classifier-free guidance that enables text-to-video  \cite{videoimagen}, and latent diffusion \cite{latent_video,text2videozero,latentvideo2,video_projected_latent,latent_flow_video,dreampose} for sampling efficiency. New techniques have also been brought up to ensure the consistency among time frames, such as additional guidance information \cite{video_diffusion,follow_pose}. This extension of TTI to text-to-video generation holds great potential for enhancing the field of digital art creation. However, it is important to note that text-to-video generation requires more computation time compared to TTI generation, as a video frame typically comprises numerous image frames to achieve smooth animation.

In addition to their application in the field oftext-to-video, TTI models can also be extended to facilitate 3D generation using either Generative Adversarial Networks (GANs) \cite{next3d,graf,3DGAN,pointdiffusion3D} or Diffusion methods \cite{lion3D,Control3Diff,rodin,3D-LDM,TriplaneDiffusion,neural-wavelet,meshdiffusion,view3D,difffacto}. The aim of 3D generation through diffusion techniques is to approximate the distribution of a 3D object's shape by utilizing random Gaussian points within 3D space. This approach enables the efficient creation of objects for 3D printing based on user descriptions, even if the desired object is not already present in an existing database. Consequently, the need for time-consuming manual design and drawing processes is eliminated, allowing for a more streamlined and expedited object creation process.

Despite the exciting prospects of TTI-based 3D generation, there are several challenges to overcome. One significant challenge is the complexity of representing three-dimensional shapes accurately \cite{wang2022art}. While TTI models have achieved impressive results in generating 2D images, extending their capabilities to 3D objects requires handling additional spatial and structural information. Ensuring the fidelity and coherence of the generated 3D shapes remains an ongoing research area, with active investigations into novel model architectures and loss functions that can effectively capture the intricacies of 3D geometry.
\section{Conclusion}
In our survey, we delve deeply into the world of TTI generation models, presenting a meticulous review of their diverse types. Our investigation encompasses their origins, highlighting the foundational models, intricate algorithms, and innovative architectures that underlie their remarkable capabilities. We assess their performance, conducting both quantitative and qualitative comparisons that shed light on their respective strengths and weaknesses.
Within the context of the ever-evolving era of large models, we endeavor to unravel the intricate ways that TTI models interact with and complement the burgeoning domain of large models. By doing so, we lay the groundwork that has made important contributions to the field of TTI generation, seeking for potential improvements that could unleash the full potential of TTI models and harnessing their power to new heights.
Moreover, our survey extends beyond the present, exploring the exciting possibilities that lie ahead for TTI models. We gaze into the horizon of future applications, envisioning novel ways they can contribute to various fields, from creative arts to data augmentation and beyond. Additionally, we examine the potential expansion of TTI models into other domains in the realm of AIGC that could change the way we retrieve information.

\section*{Acknowledgment}

DAC is funded by an NIHR Research Professorship, an RAEng Research Chair, the NIHR Oxford Biomedical Research Centre, the InnoHK Centre for Cerebro-cardiovascular Engineering, the Wellcome Trust under grant 217650/Z/19/Z, and the Oxford Pandemic Sciences Institute.
\bibliographystyle{IEEETran}
\bibliography{references}

\end{document}